\documentclass[runningheads]{llncs}
\usepackage[mobile,year=202]{eccv}

\usepackage{multirow} 
\usepackage{xcolor,colortbl}
\usepackage[table]{xcolor}
\usepackage{amssymb}
\usepackage{pifont}
\usepackage{makecell}
\usepackage{float}
\usepackage{mwe}
\usepackage{booktabs}
\usepackage{xspace}
\usepackage{footnote}
\usepackage{tabularx}

\usepackage{tikz}
\newcommand{\cornerarrow}{~\tikz[baseline=-0.6ex]{\draw[->, line width=0.6pt] (0,0) -| (0.15,-0.15);}}

\usepackage{array}

\newcolumntype{H}{%
  >{\setbox0=\hbox\bgroup}%
  c%
  <{\egroup}@{}%
}
\usepackage{hyperref}

\title{GeM\raisebox{0.05em}{\includegraphics[height=0.5em]{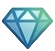}}NR: Geometry-Aware Multi-View Editing for Nonrigid Scene Changes}

\authorrunning{J. Bengtson, Y. Lochman, F. Kahl}
\titlerunning{GeM-NR: Geometry-Aware Multi-View Editing for Nonrigid Scene Changes}

\author{
Josef Bengtson$^{*}$
\hspace{0.1in}
Yaroslava Lochman$^{*}$
\hspace{0.1in}
Fredrik Kahl
}
\institute{
Chalmers University of Technology\\
\vspace{1.5em}
\texttt{
https://gem-nr.github.io
}
\vspace{-1em}
}

\begin{document}
\renewcommand{\thefootnote}{*}
\footnotetext{Equal contribution}

\maketitle
\begin{abstract}
Recent developments in multi-view image editing with generative models have brought us a step closer toward general 3D content generation and customization. Most existing works focus on \textit{rigid} or \textit{appearance-only} edits by utilizing the geometry of the unedited scene. This naturally limits these methods to edits that preserve the underlying scene structure. Current nonrigid approaches are limited to object removal and insertion, reflecting the data they are trained on. General nonrigid edits, \textit{i.e.}, edits that substantially and arbitrarily change the scene geometry, remain challenging for existing methods. We propose GeM-NR, a fast and flexible training-free approach for \textit{general} multi-view consistent image editing, including edits that drastically change the geometry and appearance of the scene. Given an anchor image edited with a chosen 2D editor and a query unedited image, GeM-NR edits the query image consistently with the anchor edit. The method incorporates multiple stages: (i) depth map estimation, where we propose a strategy to maximize the alignment between the 3D point clouds of the edited and unedited scenes, (ii) projection onto a query viewpoint, and (iii) refinement of the obtained image conditioned on the unedited query. We demonstrate the ability of our method to handle edits with significant changes in geometry and appearance, something that existing methods struggle with. We perform an extensive evaluation showing that GeM-NR improves consistency for a wide variety of edit tasks, including generating 3D representations of the edited scene. Both quantitative and qualitative results indicate the state-of-the-art performance of our method in terms of edit quality as well as geometric and photometric consistency across multiple views.
\end{abstract}
\begin{figure}
    \centering
    \caption{\textbf{The edited images produced by GeM-NR.} We handle various types of edits including significant changes in scene geometry. The edits are both photometrically and geometrically consistent across the views.}
    \includegraphics[width=\linewidth]{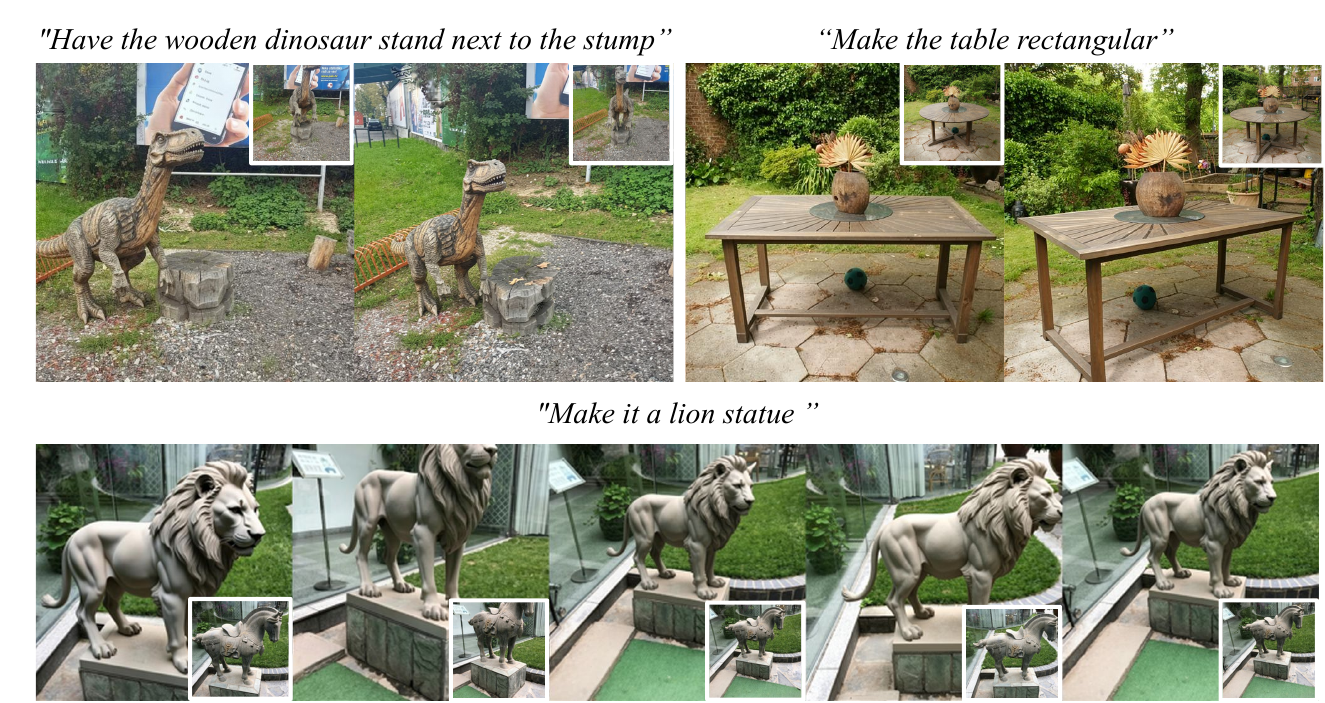}
    \label{fig:intro_fig}
\end{figure}
\section{Introduction}
\label{sec:intro}
Consistent multi-view image editing is a core capability for 3D editing, with numerous possible applications. A major challenge is in handling nonrigid edits, which drastically change the geometry of the scene. One reason for difficulties with these types of edits is that the geometry of the original unedited scene no longer holds for the edited scene. Another reason is that the more drastic edits often lead to significant inconsistencies across different views, making the task of enforcing consistency more difficult. One approach for solving this is to train a method using paired multi-view images with such edits \cite{omni3dedit}. However, limitations in the availability of paired data restrict these methods to specific types of edits, such as object insertion and removal.

One solution to the issue with significant inconsistencies in the edits is to warp an edited image into a target view and use that as conditioning when editing an image at that view, providing clear guidance on how that view should look to be consistent with the previous edit. Existing works \cite{editsplat,vica_nerf} use depth maps acquired from a 3D representation of the original scene, limiting the ability to handle nonrigid edits. In contrast, we propose estimating the scene geometry from
an edited view of the scene using a monocular depth estimator \cite{DA3}, since these depths will then be valid also for nonrigid edits. We formulate this problem as the task of performing consistent editing of a number of images of the same scene, given an initial edited image. This problem could be solved by performing single-image novel view synthesis from the initial edited image by rendering new views at the poses of the given views. One approach is to estimate geometry from the initial image and use this to warp the initial image into the target views. The problem with doing this in the setting of multi-view editing is that this does not take into account the information in the additional views of the scene. We instead propose a way to include this warping in the editing process, so that both the unedited views and the warping of the initial edit are taken into account.

Recent developments in editing methods allow for flexible and multi-reference editing \cite{qwen,flux2,flux2klein}, which could be used for multi-view editing. It is possible to condition these methods on several views and task these methods with performing multi-view consistent editing, but this is unstable and often leads to failed edits. Our contribution is that by conditioning these multi-reference methods with both an unedited image and the warping of an existing edit, it is possible to reliably perform multi-view consistent editing.

While we rely on foundation 3D models for reconstruction and multi-reference image generation models for editing, we discover novel ways of leveraging their power. In particular, our first finding is that the unedited and edited scenes together can be cast as one dynamic scene, and hence can be handled by a dynamic scene reconstruction model such as Depth Anything 3. Our second finding is that the unedited image together with its partially filled edited version is a better input for the multi-reference image generation model such as FLUX.2 than a fully edited image from another viewpoint. It has been shown that image generation models have limited ability in performing geometric tasks. However, if provided with the right geometry, they succeed at preserving it.

In summary, for an initial edit done by a preferred backbone editor, GeM-NR is able to perform multi-view consistent editing with respect to this initial edit. GeM-NR is flexible and can handle widely varied and drastic edits, including significant changes to geometry, as seen in Figure~\ref{fig:intro_fig}. It does not require any per-scene optimization and the runtime is only limited by the time it takes to perform monocular depth estimation and multi-reference editing, which for the methods we choose can be done in a few seconds per image. Our contributions can be summarized as follows:
\begin{itemize}
    \item We propose a flexible pipeline for multi-view image editing based on reconstructing the geometry of the edited scene;
    \item We discover and combine novel ways of leveraging the power of foundation models, in particular dynamic reconstruction models and multi-reference image editing models, to maximize the editing quality \emph{together} with multi-view consistency;
    \item We conduct extensive evaluation showing that GeM-NR can handle edits significantly changing geometry and appearance. We propose a more detailed evaluation pipeline integrating epipolar geometry evaluation, showing improved consistency across a variety of different editing tasks and image editing methods;
    \item Our multi-view pipeline is very fast, $2$-$4\times$ faster than the state of the art. This allows us to generate edited 3D Gaussians in less than a minute for sparse scenes.

\end{itemize}

\section{Related Work}
\label{sec:related_work}
\paragraph{Image editing using generative models.}
Diffusion models trained on vast amounts of data have led to significant advancements in both image generation \cite{diffusion,stable_diffusion,dalle2,imagen} and editing \cite{sdedit,prompt2prompt,instructpix2pix,dream_booth}. This has enabled different types of editing tasks, including instruction-based editing \cite{instructpix2pix,magic_brush,mgie}, image inpainting \cite{brushnet,PowerPaint,repaint}, style transfer \cite{inst,style_aligned,style_injection,art_adapter}, and multi-reference editing \cite{omnigen,qwen,flux2,flux2klein}. Recent developments in flow matching \cite{flow-matching,rectified_flow} and multi-modal transformer architectures \cite{mmdit} have led to high-quality unified editing models \cite{flux1,qwen,flux2,flux2klein} that can reliably perform significant and precise edits. Our work focuses on using these powerful 2D image editing methods to perform consistent multi-view editing.

\paragraph{Multi-view consistent image editing.}
Much of the research on multi-view consistent editing focuses on adapting 2D image editing models to edit an existing 3D representation, such as NeRFs or 3D Gaussians. Instruct-NeRF2NeRF \cite{haque2023instruct} introduces iterative dataset update (IDU), which achieves consistent edits by iteratively editing views and updating the 3D representation. This strategy is further adopted in several works \cite{chen2023gaussianeditor,vcedit,mirzaei2023watchyoursteps,ProteusNeRF,vica_nerf}. Other approaches leverage the 3D Gaussian representation of the unedited scene, e.g., by using its geometry to guide the editing process \cite{gaussctrl,editsplat,core_editor,diffusion_feature_field}, or by rendering smooth sequences and using video diffusion priors to maintain 3D consistency 
%to enable taking inspiration from video editing methods 
\cite{dge,vip3de,qu2025editcast3dsingleframeguided3dediting}. A related technique is to iteratively update the 3D representation with edited images, thereby inheriting consistency from the underlying representation \cite{dge,editsplat,gaussctrl,vcedit,vip3de,Dynamic-eDiTor}. These approaches require dense views to obtain a high-quality 3D representation of the unedited scene. An alternative that avoids this requirement is to use correspondences between the unedited images to guide the editing such that corresponding points are edited consistently \cite{efficient-nerf2nerf,edicho,correspondence_guidance}. While directly encouraging multi-view consistency, it has limitations for edits that significantly change the geometry of the scene, since the correspondences from the unedited images do not hold after the edit. Another line of work directly edits a 3D representation or a set of multi-view images, but requires paired 3D data, limiting most methods to single objects and synthetic scenes \cite{xia2025scalableconsistent3dediting,ye2025nano3dtrainingfreeapproachefficient,li2025voxhammer,ava-nvs}. Recent works try to extend this approach to real images, either by using a pipeline based on developments in 2D image editing and VLMs \cite{gemini2_5,qwen2} to create paired multi-view edits for specific edit types \cite{omni3dedit} or by performing reinforcement-learning-based finetuning and leveraging the 3D foundation model VGGT \cite{vggt} as a consistency reward \cite{rl3dedit}. In contrast, GeM-NR is training-free and does not require dense views: we estimate the geometry of the edited scene from the initial edit and use it to compute a warping that conditions the editing of the second view, propagating the edit consistently even for nonrigid scene changes.

\begin{figure}[t]
    \centering
    \caption{\textbf{Method overview.} GeM-NR estimates the geometry of the edited scene while maintaining global alignment with the unedited scene, projects the resulting point cloud onto the target view, and uses the resulting warp to condition the editing of the next image.}
    \includegraphics[width=\linewidth]{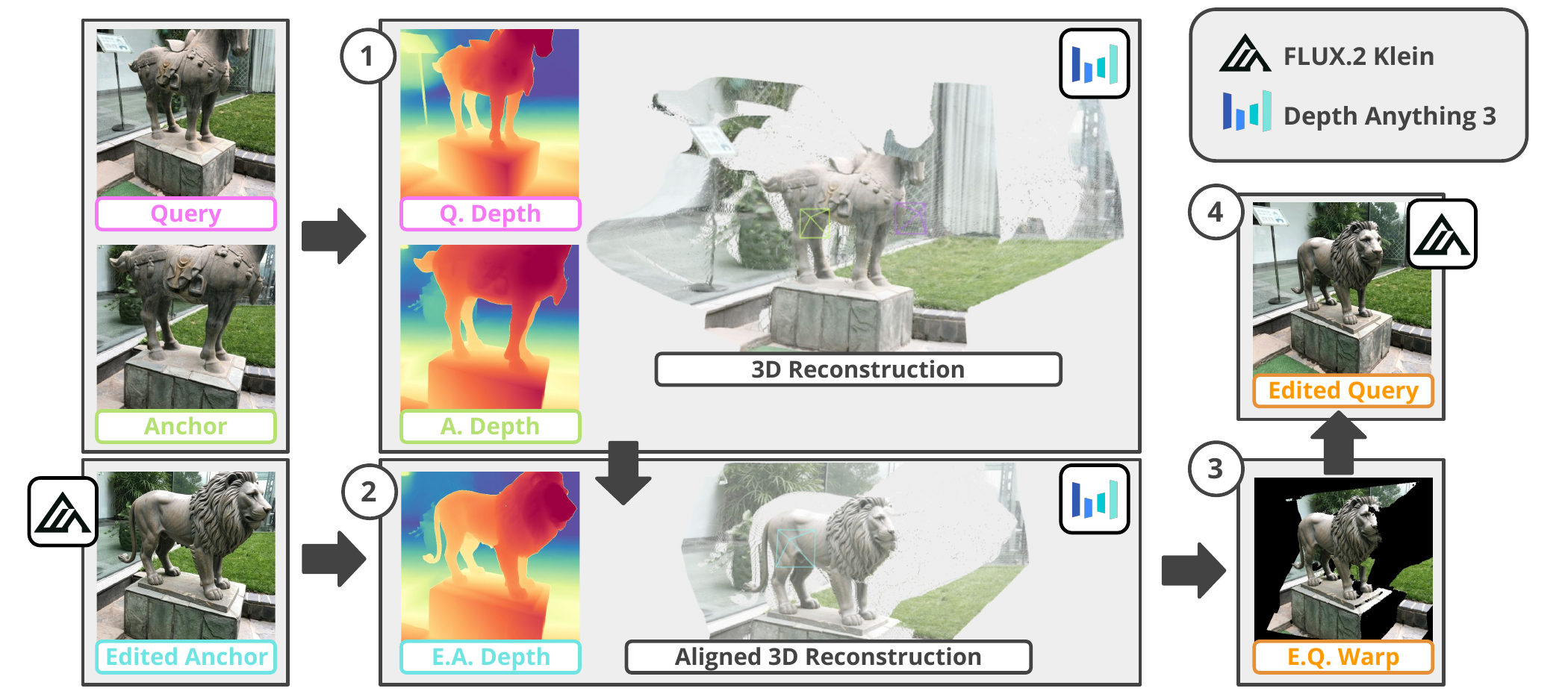}
    \vspace{-2em}
    \label{fig:method}
\end{figure}

\paragraph{Warping-based conditioning.}
One way to improve multi-view consistency is to warp an input view to a target viewpoint. This has been used for single-image novel view synthesis, where an input image is warped to a target pose, conditioning the generation for the corresponding view \cite{megascenes,infinite_nature,viewcrafter,multidiff,gen_warp,nvs-solver,wave}. Several works also use warping to improve consistency in 3D editing, either by leveraging depth from the unedited 3D representation to blend edits across views \cite{editsplat,vica_nerf}, conditioning the editing on estimated depths \cite{gaussctrl,date_nerf}, or warping attention feature maps \cite{attention-warp}. These approaches are primarily designed for rigid edits that roughly preserve the scene geometry, since the depth is derived from the unedited scene. In contrast, we propose to utilize the geometry of the edited scene, making the approach applicable to nonrigid scene changes. Recent geometry estimation models \cite{dpt,midas,DA,DA3} can be leveraged to obtain the geometry of the edited scene from the initial edited image, and to warp the image into a target view. Our approach adapts this for nonrigid multi-view editing by combining the warped edit with the image to be edited to preserve multi-view details and consistency even when the edit changes scene geometry.

\section{Method}
\label{sec:method}
\subsection{Problem formulation}
Given $N$ source views $\{I^1_{src}, I^2_{src},...,I^N_{src}\}$ and a text description $T$ of the desired edit, the goal is to edit the images, obtaining $\{I^1_{edited}, I^2_{edited},...,I^N_{edited}\}$, such that their appearance and geometry are consistent across views, while remaining faithful to the edit description $T$. One way to reformulate the problem is to first choose an anchor image $A_{src}$ that is to be edited independently: $A_{edited} = f(A_{src}, T)$, where $f$ is a preferred image editing model. The remaining images, which we refer to as query images and denote by $\{Q^i_{src}\}$, can then be edited conditioned on $A_{edited}$ for consistent editing. Thus, we simplify the task to processing triplets consecutively: given $\{Q_{src}, A_{src}, A_{edited}\}$, the goal is to produce $Q_{edited}$ that is consistent with $A_{edited}$. As of now, existing image editing models do not guarantee that independently edited images, that is, $Q_{edited} = f(Q_{src}, T)$, will retain such consistency.

\begin{figure}[t]
\centering
\caption{\textbf{Monocular vs. our joint depth estimation.} In regions where the geometry is unchanged by the edit (zoomed-in), the depths of the unedited and edited scene from the same viewpoint should coincide. Monocular depths are not as accurately aligned w.r.t the unedited scene as our jointly estimated depths, which use all unedited and edited images.}
\includegraphics[width=\textwidth]{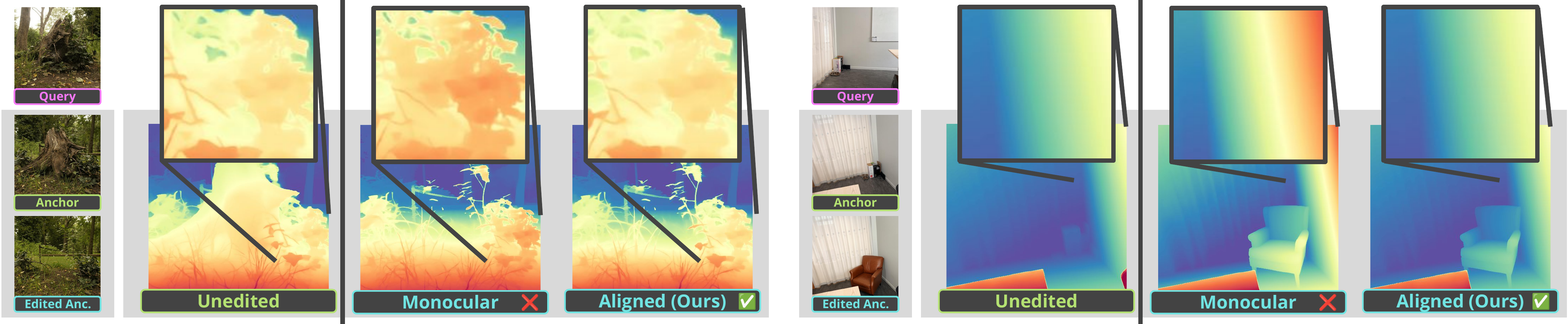}
\vspace{-2em}
\label{fig:monocular_depth}
\end{figure}

\subsection{Overview}
Our editing pipeline should be able to handle significant changes in the scene. For appearance changes, since at least two unedited views are available, one could use the geometry of the unedited scene obtained with a 3D reconstruction pipeline. For challenging geometric changes, however, we cannot utilize the geometry acquired from the unedited images. Instead, we propose to use the geometry of the edited scene. Moreover, we need to understand how the geometry of the edited scene relates to that of the unedited scene. In other words, the two 3D representations need to be aligned. However, solving the alignment accurately and robustly is challenging, especially if the majority of the scene geometry is changed. We sidestep the alignment problem altogether and instead aim to jointly estimate the 3D representations of both the edited and unedited scene as realizations of one dynamic scene, which automatically results in a global alignment. This becomes possible with the recent advances in dynamic scene reconstruction~\cite{DA3}. The 3D representation of the globally aligned edited scene can now be used to render an image at a query viewpoint. Naturally, at this viewpoint, some regions need to be filled in as they were not seen from the anchor viewpoint. A source query image provides additional information about what should be depicted in these areas. A multi-reference image generation model~\cite{flux2, flux2klein} can therefore be used to fulfill the task of final edit generation given an unedited image and its partially edited version obtained by warping. Figure~\ref{fig:method} presents an overview of our approach. Since the overall pipeline only requires one forward pass and utilizes efficient foundation models, it takes only a few seconds to edit a pair of images consistently.

Having a method that can edit two images consistently, we extend the approach to editing sets of images by repeating the process for all the remaining query images. We can then pass the resulting set of consistently edited images through a feedforward method such as AnySplat \cite{anysplat} to obtain a 3D Gaussian Splatting (3DGS) representation of the edited scene.

\begin{figure}[t]
  \centering
  \caption{\textbf{Qualitative comparison of query-anchor consistency with Qwen anchor backbone}. Our method GeM-NR preserves photometric and geometric details of the anchor edit.}
 {\scriptsize \textit{``Remove the bear statue''}}\\[1pt]
  \includegraphics[width=\textwidth]{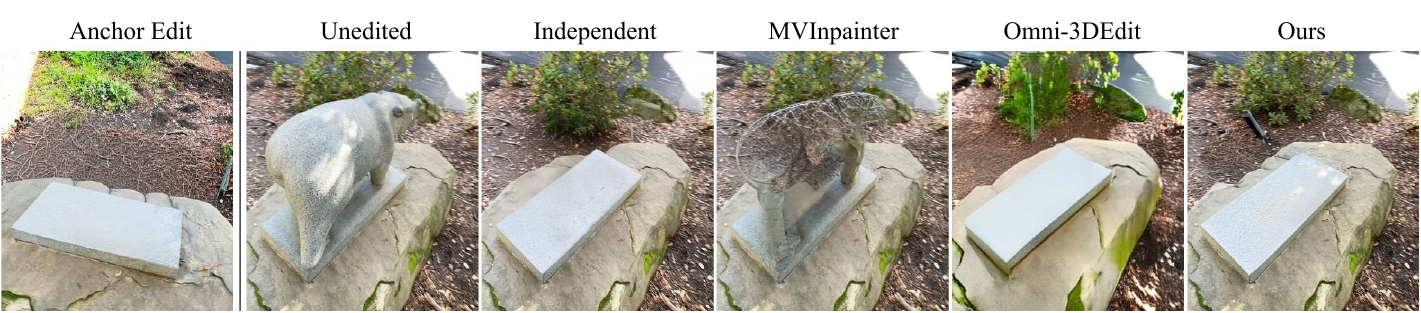} \\
    {\scriptsize \textit{``Change bear statue to a sitting dog''}}\\[1pt]
  \includegraphics[width=\textwidth]{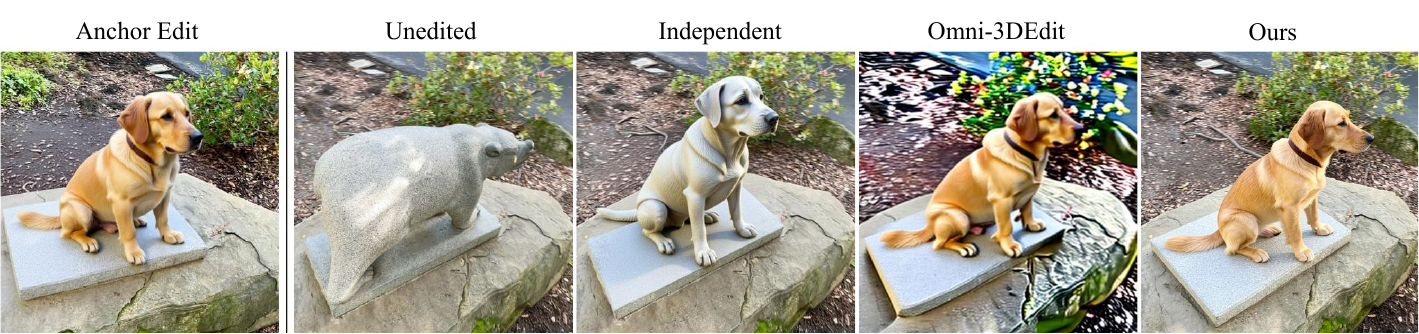}\\
   {\scriptsize \textit{``Add large colorful shiny gemstones to the statue''}}\\[1pt]
  \includegraphics[width=\textwidth]{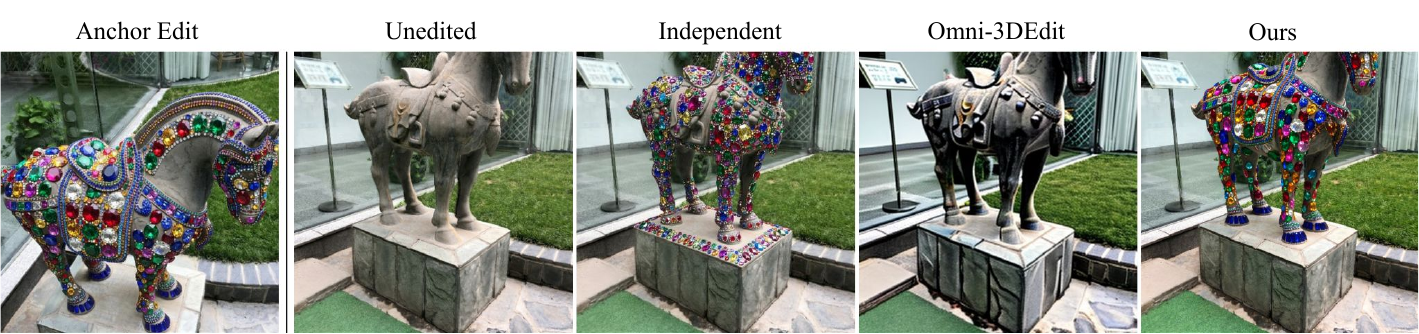} 
   
  \label{fig:qual_results_mv_single_row}
\end{figure}

\subsubsection{Edited scene geometry estimation}
We first pass $A_{src}$ and $Q_{src}$ through the reconstruction model Depth Anything 3~\cite{DA3} to obtain camera intrinsics, $K_A$ and $K_Q$, respectively, and extrinsics, $P_A$ and $P^0_Q$. The pose $P^0_Q$ is then refined into $P_Q$ with the classic relative pose estimation from two views using RoMa~\cite{roma} matches. The refinement affects the quality of the subsequent joint reconstruction discussed below. Further, we show in Section~\ref{sec:ablation} that it improves the quality of the final edits.

Next, the three images $\{A_{src}, A_{edited}, Q_{src}\}$ and the corresponding camera intrinsics and extrinsics $\{(K_A, P_A), (K_{EA}, P_{EA}), (K_Q, P_Q)\}$ are passed through Depth Anything 3, constraining the output cameras to be equal to the ones provided. We use the same camera parameters for $A_{edited}$ as for $A_{src}$: $K_{EA}=K_A$, $P_{EA}=P_A$. This is based on the assumption that the edit does not change the global motion. E.g., if the edit is: ``Change the view to the other side of the corridor'' or ``Turn to the right and show what is there'', it violates our assumption. However, if the edit is: ``Turn [depicted object] around'', it aligns well with our assumption as it keeps the global scene pose unchanged. See also an example with the edit of the similar form, namely ``Have the wooden dinosaur stand next to the stump'', in Figure~\ref{fig:intro_fig}. Depth Anything 3 is trained to tackle dynamic scene reconstruction, and we leverage this ability for challenging nonrigid edits. The model views $\{A_{src}, A_{edited}, Q_{src}\}$ as images of a single dynamic scene, where possible changes in geometry and photometry over time appear at $A_{edited}$.

\subsubsection{Edit initialization with warping}

A partially-filled edited image is rendered at a query viewpoint by projecting the point cloud of the edited scene, which can also be seen as warping $A_{edited}$, giving $Q_{warp}$. The warping provides dense guidance on roughly how the edited views should be to ensure consistency with the edited anchor view. 

In this work, we use depth map / point cloud representation for both estimation and rendering. In theory, any other representation, such as NeRFs or 3D Gaussians, could work as long as such a representation can be faithfully obtained from only a few images, in our case as few as three images, and the underlying estimation method supports dynamic scenes. Another option is to upgrade to a desired representation starting from a point cloud. A challenge with this approach is that it has access to very limited data, i.e., a single edited image with the corresponding depth map. At this point, we found that manipulating point clouds directly works best.

\paragraph{Why not use monocular depth?}
A slightly simpler approach could be to use monocular depths estimated from an edited anchor image $A_{edited}$ to warp this image from camera $(K_A, P_A)$ to camera $(K_Q, P_Q)$. In Figure~\ref{fig:monocular_depth}, we show qualitative examples of obtaining a monocular depth map from a calibrated edited image as compared to our alignment approach. Ideally, the depth values of $A_{src}$ and $A_{edited}$ should coincide in regions where the geometry has not been changed. In the examples shown, we zoom in on such regions (leaves in the left example and wall in the right example). Monocular depth estimates (even when calibrated) do not follow the multi-view depths of the unedited scene as closely as our approach, in which the depth maps of all images are estimated jointly.

\subsubsection{Edit refinement with multi-reference editing}
To fill in the gaps in $Q_{warp}$, one could use an image inpainting model. The problem with this approach is that it does not guarantee the consistency of the filled-in areas with those in $Q_{src}$. We need to be able to constrain the image generation process both by forcing the edit to closely follow $Q_{warp}$ where possible, and also by letting the model infer how to fill the remaining areas based on the semantics of the corresponding regions in $Q_{src}$ and the overall content in $Q_{warp}$. Hence, the model should be able to accept multiple visual inputs and understand the relation between them from the text description, which we refer to as multi-reference image editing. 

Recent image editing methods \cite{qwen,flux2,flux2klein} allow for multi-reference inputs: a list of images $I_{list}=[I_1,I_2,...]$ and a text prompt $\hat T$ specifying how the images should be combined to create a final image $I_{edited} = f_{multi}\left(I_{list},\hat T\right)$. 
We propose to include the warping of the initial edited view into the target view as conditioning when editing. This constraint is provided directly in the image frame of the view that should be edited, making it easier for the model to create an image that respects the warping from the initial edited view, encouraging consistency, while also respecting the additional information given in the unedited target view:
\begin{equation}
    Q_{edited} = f_{multi}\left([Q_{src}, Q_{warp}], \text{concat}(T, T_+)\right),
\end{equation}
where our additional instruction $T_+$ is: ``The suggested appearance is in the second image. Stick to this change, but refine it to keep consistency with respect to the first image.''

\begin{figure}[t]
\centering
\caption{\textbf{Application of our method in interior design.}}
\includegraphics[width=\textwidth]{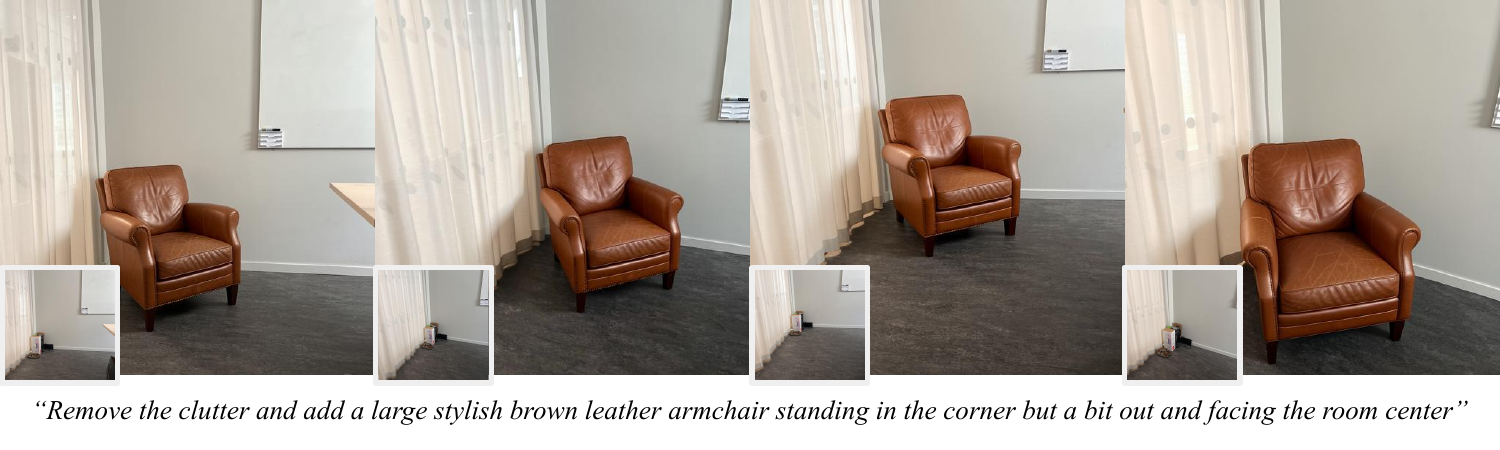}
\label{fig:armchair}
\end{figure}

\section{Experiments}
\label{sec:experiments}
\begin{figure}[t]
  \centering
  \caption{\textbf{Multi-view editing of the 10 image-sequences with Qwen anchor backbone}. GeM-NR successfully edits the full sequence while maintaining multi-view consistency.}
  {\scriptsize \textit{``Change the bicycle into an e-scooter''}}\\
  \includegraphics[width=0.9\textwidth]{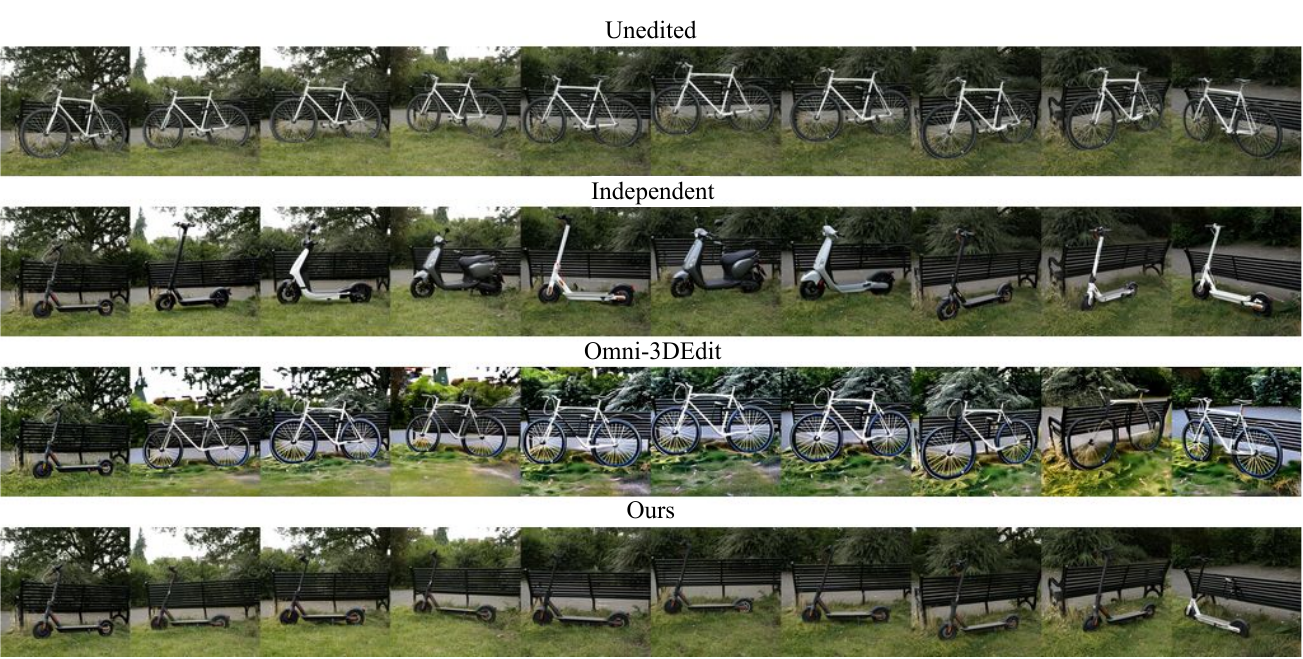}\\[0pt]
  {\scriptsize \textit{``Make him wear a detailed fancy suit''}}\\
  \includegraphics[width=0.9\textwidth]{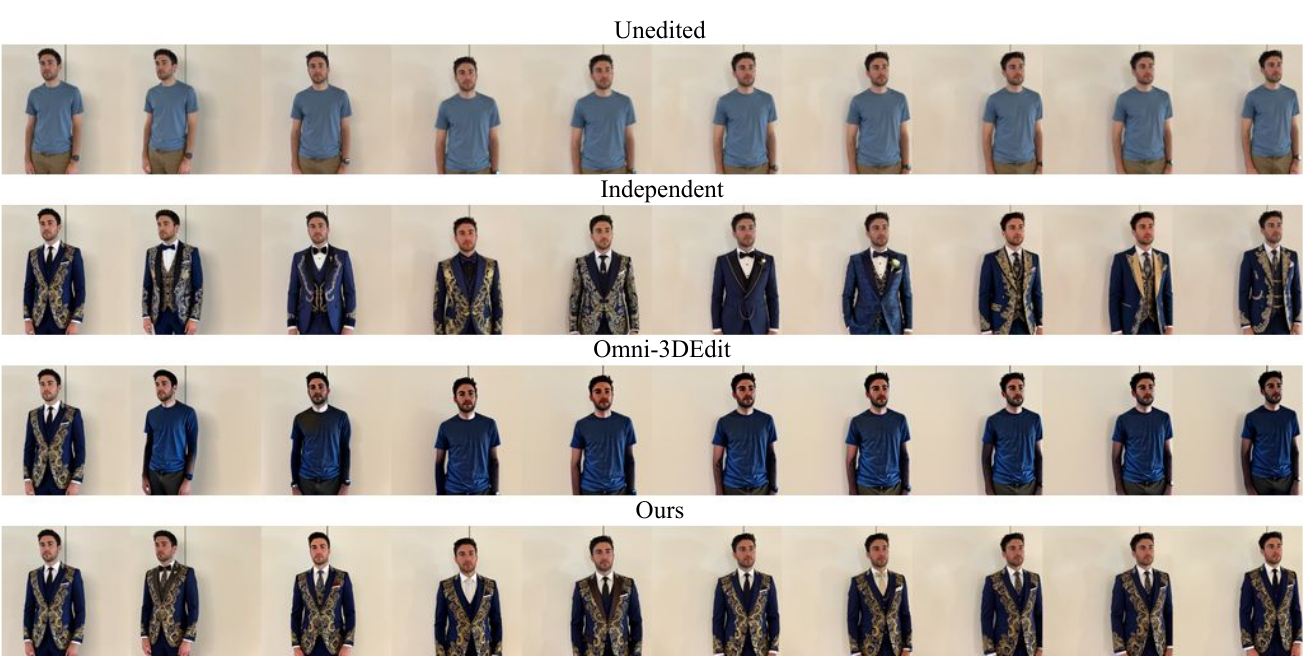}
  \label{fig:qual_results_multi-view}
\end{figure}

\begin{figure}[t]
  \centering
  \caption{\textbf{Renders from 3DGS of edited scene}. Our approach gives sharp renders that preserve the details from the anchor edit.
  }
  \label{fig:qual_results_renders}
  \includegraphics[width=\textwidth]{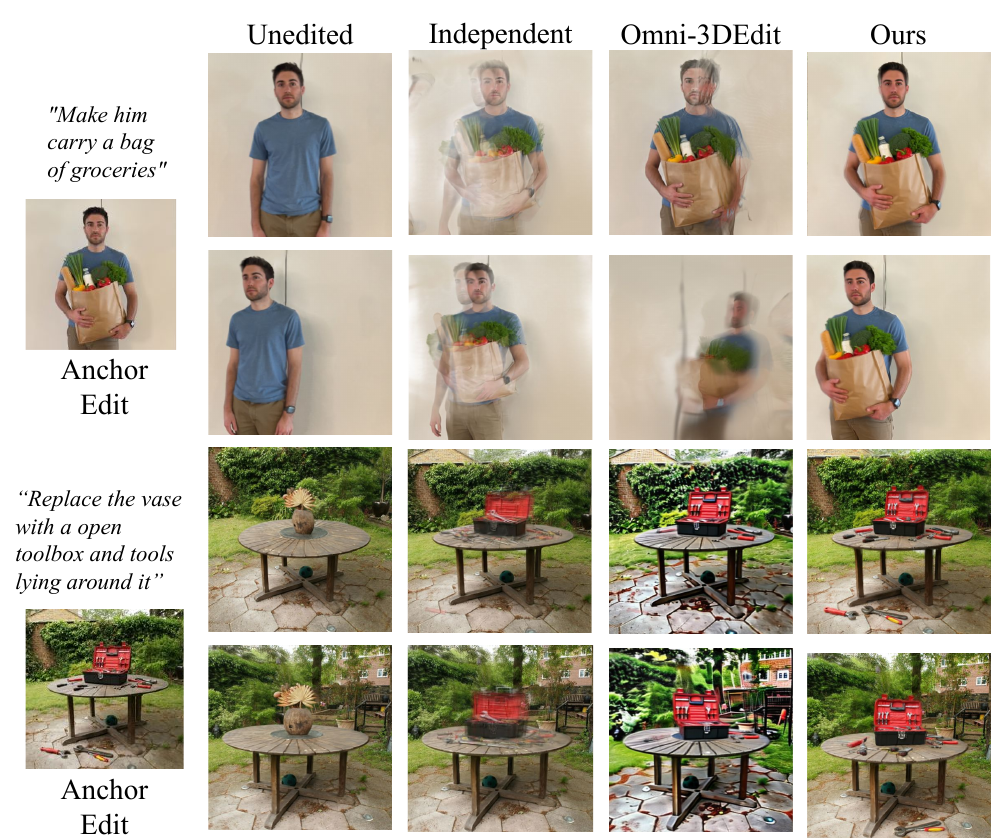}
\end{figure}

\begin{figure}[t]
  \centering
  \caption{\textbf{Qualitative image pair editing examples} for Edicho and our GeM-NR.}
  \includegraphics[width=\textwidth]{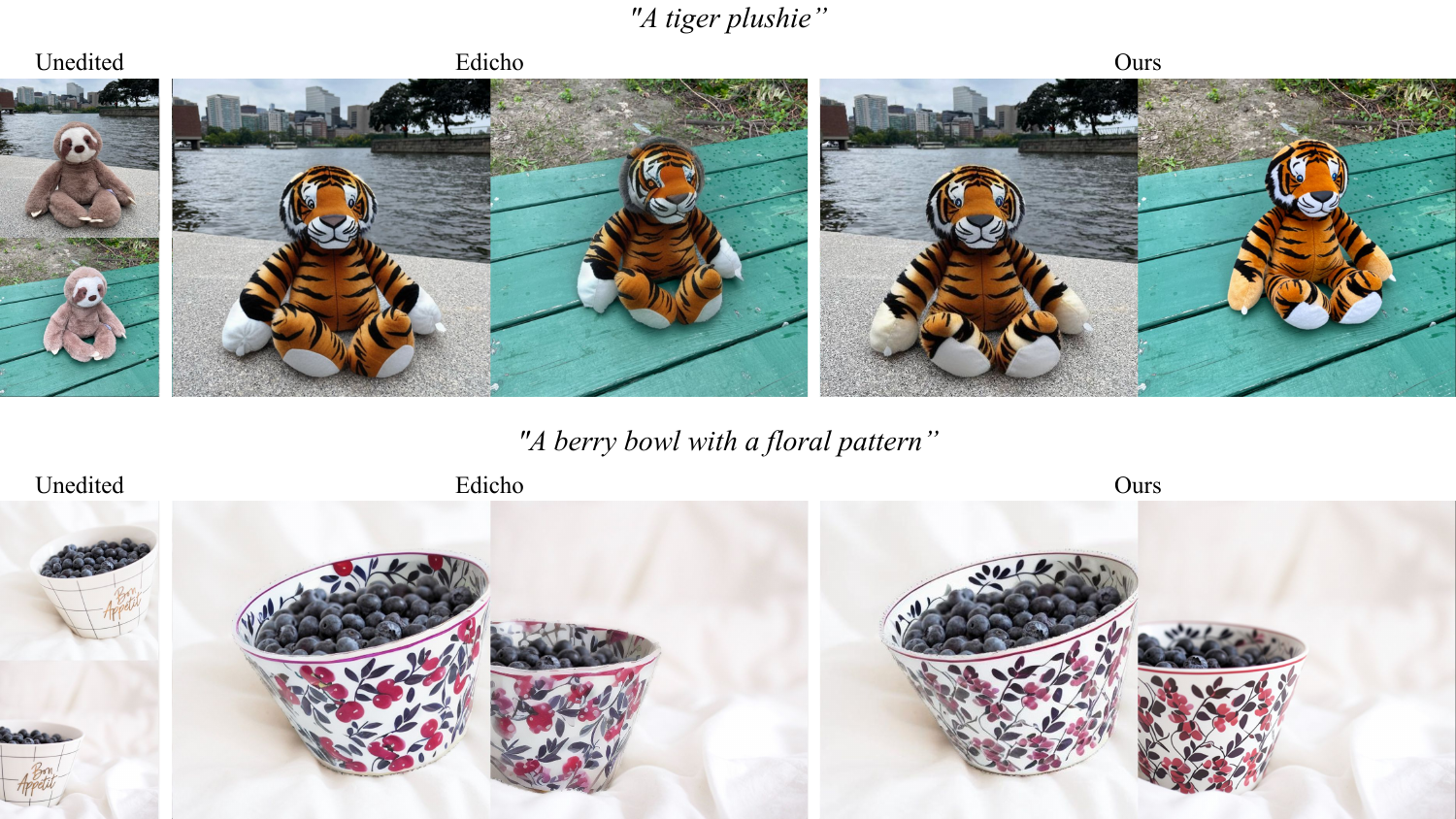}
  \label{fig:qual_results_pairs}
\end{figure}

We divide our evaluation into two main tasks: general multi-view editing (Figures~\ref{fig:armchair}-\ref{fig:qual_results_renders}) and image pair editing (Figure~\ref{fig:qual_results_pairs}). For general multi-view editing, we also evaluate edited 3DGS representations (Figure~\ref{fig:qual_results_renders}). We evaluate view consistency for both tasks. Additional experimental results can be found in Section~A in the supplementary material.

\subsection{Experimental setup}
We evaluate and compare GeM-NR across a variety of different edit types, on a single NVIDIA A100 with 80GB. Our evaluation data includes 38 prompts over 4 editing categories: general nonrigid edits, object addition, object removal, and appearance change. We use a combination of 15 test scenes taken from the datasets SPIn-NeRF \cite{spinnerf}, IN2N \cite{haque2023instruct},  Mip-NeRF 360 \cite{barron2022mipnerf360}, and BlendedMVS \cite{blendedmvs}. We also report ablation study results on 4 held-out validation scenes, using 11 prompts. For this task, we compare with Omni-3DEdit \cite{omni3dedit} which performs multi-view editing by directly editing a set of up to 10 images given one edited anchor image. They achieve this by creating a dataset of paired edits for specific scenarios such as object removal, insertion, and appearance change. They can also handle more complex edits by first performing object removal and then object addition. Existing training-free 3DGS-based methods \cite{vip3de,dge,editsplat} require dense scene capture, limiting direct comparison on the task of sparse-view editing. Sparse-view editing methods based on consistent inpainting, such as MVInpainter \cite{mvinpainter}, do not support general nonrigid editing, so we compare GeM-NR with MVInpainter specifically on object removal, a task it does support. MVInpainter also supports object addition, but this requires manual 3D bounding box annotations, which we do not assume are available.

For evaluating consistent image pair editing, we use image pairs taken from DreamBooth~\cite{dream_booth} and Mip-NeRF 360 \cite{barron2022mipnerf360}, using a total of 38 pairs in the test set. For this task, we compare with Edicho \cite{edicho}, which performs consistent image editing by computing explicit correspondences from the unedited images that are used to guide the denoising process. They apply this approach to two different 2D editing methods, ControlNet \cite{controlnet,uni-controlnet} for global editing and BrushNet \cite{brushnet} for inpainting-based editing. Their code release only includes the inpainting-based approach \cite{brushnet}, so this is the one that we compare to.

\paragraph{Implementation details.} 
We use FLUX.2 [klein] \cite{flux2klein} as our backbone multi-reference image editing model. For geometry estimation, we use Depth Anything 3 \cite{DA3}. For the general multi-view editing, initial anchor edits are generated using Qwen \cite{qwen} (version Qwen-Image-Edit-2509) which is the editing method used when training Omni-3DEdit. When comparing image pair editing performance to Edicho, we use BrushNet \cite{brushnet} for the initial anchor edits, since this is the method used by Edicho. BrushNet requires masks and inpaints the masked object area. We adapt our method to work with this type of input by providing a masked anchor edit as a reference, and estimating the object geometry instead of the full scene. Finally, for all tasks, we also test FLUX.2 [klein] \cite{flux2klein} for the anchor edits. For generating a 3DGS representation from the edited multi-view images, we use AnySplat \cite{anysplat}, a feedforward method that in seconds can generate a 3DGS from a set of sparse unposed images.

\begin{table}[]
\centering
\caption{\textbf{Consistency evaluation of edited multi-view images}. GeM-NR improves multi-view consistency and preserves edit instructions, cf.\ high TA scores.}
\scriptsize
\resizebox{\textwidth}{!}{%
\begin{tabular}{p{0.2cm}llcccccc}
\toprule
 & \textbf{} & \textbf{} & \multicolumn{2}{c}{\textbf{\scriptsize\makecell{Geometric\\Edit Consistency}}} & \multicolumn{3}{c}{\textbf{\scriptsize\makecell{3D Reconstruction\\Consistency}}} & \textbf{\scriptsize\makecell{Text\\Alignment}}\\
 & & & MEt3R$\downarrow$ & mAA, \% $\uparrow$ & PSNR$\uparrow$ & SSIM$\uparrow$ & LPIPS$\downarrow$ & TA / TA dir $\uparrow$\\
\midrule %%%%%%%%%%%%%%%%%%%%%%%%%%%%%%%
\multicolumn{3}{l}{\textit{\scriptsize All Edit Types}\cornerarrow}\\
\midrule %%%%%%%%%%%%%%%%%%%%%%%%%%%%%%%
& \multirow{3}{*}{Qwen}
   & Independent   & 0.248 & 31.47 & 20.70 & 0.677 & 0.213 & \textbf{0.258} / \textbf{0.202}\\
 & & Omni-3DEdit   & 0.236 & 33.99 & 20.06 & \textbf{0.690} & 0.164 & 0.230 / 0.127 \\
 & & \textbf{GeM-NR} & 0.194 & 36.71 & 21.63 & 0.653 & 0.167 & 0.252 / 0.198\\
 \cmidrule{2-9}
 & \multirow{3}{*}{FLUX.2}
   & Independent   & 0.231 & 34.03 & 21.12 & 0.670 & 0.203 & 0.254 / 0.190\\
 & & Omni-3DEdit   & 0.238 & 34.27  & 19.57 & 0.677 & 0.171 & 0.233 / 0.122\\
 & & \textbf{GeM-NR} & \textbf{0.190} & \textbf{38.58}   & \textbf{21.96} & 0.659 & \textbf{0.162} & 0.254 / 0.193\\
 \cmidrule{2-9}
  & & \textit{Unedited} & 0.186 & 47.71  & 25.76 & 0.769 & 0.116 & 0.204 / -\\
\midrule %%%%%%%%%%%%%%%%%%%%%%%%%%%%%%%
\multicolumn{3}{l}{\textit{\scriptsize General Nonrigid}\cornerarrow}\\
\midrule %%%%%%%%%%%%%%%%%%%%%%%%%%%%%%%
 & \multirow{3}{*}{Qwen}
   & Independent   &  0.303 & 23.15 & 18.62 &0.613&  0.284  & 0.265 / 0.240\\
 & & Omni-3DEdit   & 0.288 & 23.78 & 18.21 & \textbf{0.662} & 0.205 & 0.251 / 0.184\\
 & & \textbf{GeM-NR} & \textbf{0.215} & 31.56 & 20.68 & 0.612 & 0.194 & 0.268 / \textbf{0.252}\\
 \cmidrule{2-9}
 & \multirow{3}{*}{FLUX.2}
   & Independent   & 0.291 & 24.91 & 18.93 & 0.598 & 0.275 & \textbf{0.269} / 0.234\\
 & & Omni-3DEdit   & 0.288 & 21.56 & 17.60 & 0.628 & 0.216  & 0.248 / 0.172\\
 & & \textbf{GeM-NR} & 0.216 & \textbf{32.35} & \textbf{20.91} & 0.607 & \textbf{0.188} & 0.268 /  0.247\\
 \cmidrule{2-9}
 & & \textit{Unedited} & 0.216 & 47.31   & 25.23 & 0.745 & 0.128&0.202 / - \\
\midrule %%%%%%%%%%%%%%%%%%%%%%%%%%%%%%%
\multicolumn{3}{l}{\textit{\scriptsize Object Addition}\cornerarrow}\\
\midrule %%%%%%%%%%%%%%%%%%%%%%%%%%%%%%%
 & \multirow{3}{*}{Qwen}
   & Independent & 0.217 & 41.22 & 20.72 & 0.738 & 0.220 & \textbf{0.242} / \textbf{0.190}\\
 & & Omni-3DEdit & 0.125 & 46.60 & 23.46 & \textbf{0.780} & \textbf{0.100} & 0.221 / 0.120 \\
 & & \textbf{GeM-NR}        & \textbf{0.120} & 46.61 & 24.40 & 0.766 & 0.105 & 0.226 / 0.151\\
 \cmidrule{2-9}
 & \multirow{3}{*}{FLUX.2}
   & Independent & 0.165 & 45.97 &22.82 & 0.769& 0.158 & 0.240 / 0.165 \\
 & & Omni-3DEdit & 0.134 &46.99& 22.92& 0.779& 0.107   & 0.232 / 0.121\\
 & & \textbf{GeM-NR}        & 0.121 & \textbf{47.17} & \textbf{24.78}& 0.777 &0.111 & 0.238 / 0.144 \\
 \cmidrule{2-9}
 & & \textit{Unedited} & 0.097 & 53.33 & 29.03 & 0.860 & 0.070 &0.191 / -\\ 
\midrule %%%%%%%%%%%%%%%%%%%%%%%%%%%%%%%
\multicolumn{3}{l}{\textit{\scriptsize Object Removal}\cornerarrow}\\
\midrule %%%%%%%%%%%%%%%%%%%%%%%%%%%%%%%
 & \multirow{3}{*}{Qwen}
   & Independent & 0.166 & 41.79 & 24.06 & 0.734 & 0.143  & 0.207 / \textbf{0.156}\\ &
    & MVInpainter & 0.177 & 43.61 & \textbf{24.85} & \textbf{0.751} & 0.134 & \textbf{0.208} / 0.143\\  
 & & Omni-3DEdit & 0.203 & 38.27 & 20.83 & 0.690 & 0.174  &          0.201  / 0.114 \\
 & & \textbf{GeM-NR}        & 0.153 & 45.66 & 24.04 & 0.712 & 0.135 & 0.201  / 0.155 \\
 \cmidrule{2-9}
 & \multirow{3}{*}{FLUX.2}
   & Independent
   & 0.165 & 44.62 & 24.66 & 0.741& 0.142                                & 0.204 /  \textbf{0.156}\\
 & & Omni-3DEdit & 0.202 & 40.90 & 20.81& 0.698& 0.170                                     & 0.199 /  0.104 \\
 & & \textbf{GeM-NR}        & \textbf{0.151} & \textbf{46.43} & 24.44 & 0.719&\textbf{0.133} & 0.204 /  0.154 \\
 \cmidrule{2-9}
 & & \textit{Unedited} & 0.152 & 48.15 & 25.95 & 0.790 & 0.101 &0.203 / -\\ 
\midrule %%%%%%%%%%%%%%%%%%%%%%%%%%%%%%%
\multicolumn{3}{l}{\textit{\scriptsize Appearance Change}\cornerarrow}\\
\midrule %%%%%%%%%%%%%%%%%%%%%%%%%%%%%%%
 & \multirow{3}{*}{Qwen}
   & Independent  & 0.256 & 29.15 &\textbf{20.74} &0.673 & 0.193& \textbf{0.278} / \textbf{0.200}\\
 & & Omni-3DEdit  & 0.257 & 34.03 &19.67 &\textbf{0.674} &\textbf{0.158} & 0.232 / 0.099\\
 & & \textbf{GeM-NR}& 0.224  & 32.87 & 20.29 & 0.614 &0.187 &0.271 / 0.199\\ \cmidrule{2-9}
 & \multirow{3}{*}{FLUX.2}
   & Independent   & 0.244 & 31.27 & 20.52& 0.652& 0.199&0.270 / 0.185\\
 & & Omni-3DEdit   & 0.260 & 34.95 & 19.09 & 0.663 & 0.168&0.237 / 0.098\\
 & & \textbf{GeM-NR} & \textbf{0.214} & \textbf{36.31} & 20.63 & 0.626 & 0.174 &0.270 / 0.193\\
 \cmidrule{2-9}
 & & \textit{Unedited} & 0.214 & 46.17 & 24.80&0.741 & 0.132 &0.210 / -\\
 \bottomrule
\end{tabular}%
}
\label{tab:MultiViewConsistency}
\end{table}

GeM-NR offers competitive runtime performance, taking on average \textbf{$\sim3.4$ seconds per image}, making it comparable to Edicho ($3.5$ seconds) for pair editing. For multi-view editing, GeM-NR is substantially faster than Omni-3DEdit, which takes $\sim8.3$ seconds per image, and the difference is even more pronounced for complex nonrigid edits, where Omni-3DEdit requires $\sim16.6$ seconds per image. This is because Omni-3DEdit requires two forward passes for nonrigid edits other than object insertion and removal.

\paragraph{Evaluation metrics.}
For consistent image pair editing, we follow \cite{edicho} and use CLIP~\cite{clip} to evaluate text alignment and edit consistency.
For general multi-view editing, we evaluate both consistency and text alignment across the edited views. Multi-view consistency is measured with MEt3R~\cite{met3r}, which compares DINO~\cite{dino} embeddings at matches obtained by Dust3R~\cite{dust3r}, and by 3D reconstruction consistency metrics that measure how well a 3DGS can be fit to the edited images, using PSNR/SSIM/LPIPS to compare against AnySplat \cite{anysplat} renders.
To verify that the edited images preserve the same relative poses as for the unedited images, we also report mean average accuracies (mAA) computed by thresholding symmetrized epipolar distances and matching confidences. Finally, we evaluate the resulting 3D Gaussians by measuring how well their renderings align with the text prompt.

\begin{figure}[t]
    \centering
    \caption{\textbf{3D reconstructions of the edited images}, obtained using VGGT \cite{vggt}.}
    \includegraphics[width=\linewidth,clip]{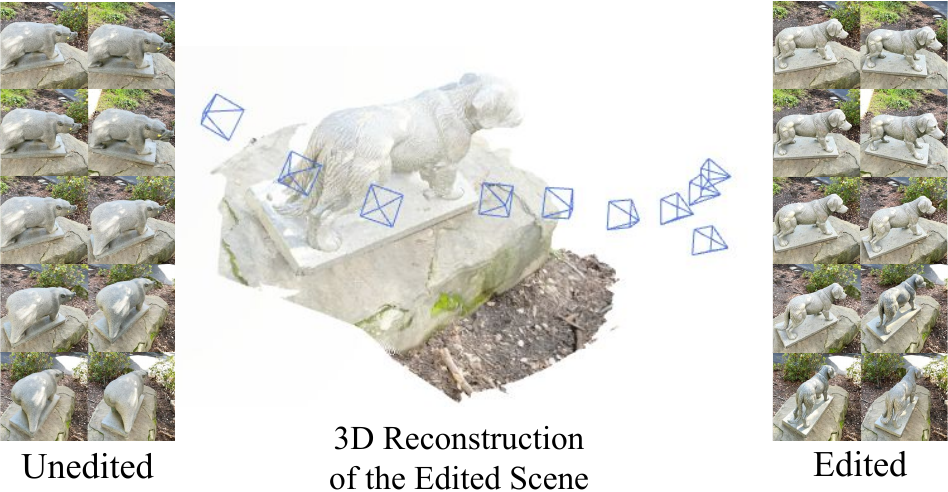}
    \label{fig:vggt}
\end{figure}

\begin{table}[h]
\centering
\footnotesize
\caption{\textbf{Text alignment (TA$\uparrow$ / TA dir$\uparrow$) of the renders from 3DGS of edited scene with Qwen anchor backbone}. GeM-NR improves over Omni-3DEdit. It especially improves for general nonrigid edits and object addition, where independent and Omni-3DEdit struggle. Comparable results are achieved when using FLUX.2 [klein] anchor backbone, as can be seen in Section~A.5 in the supplementary material.} 
\setlength{\tabcolsep}{4pt}
\resizebox{\linewidth}{!}{%
\begin{tabular}{lccccc}
\toprule
& All Edit Types & General Nonrigid & Object Addition & Object Removal & Appearance Change \\
\midrule

Independent 
& 0.250 / 0.191 
& 0.249 / 0.214 
& 0.225 / 0.134 
& \textbf{0.212} / 0.170 
& \textbf{0.275} / \textbf{0.206} \\

Omni-3DEdit 
& 0.229 / 0.130 
& 0.244 / 0.194 
& 0.232 / 0.130 
& 0.207 / 0.126 
& 0.226 / 0.092 \\

MVInpainter& -
& -
& -
&0.209 / 0.132
& - \\

\textbf{GeM-NR} 
& \textbf{0.258} / \textbf{0.212} 
& \textbf{0.269} / \textbf{0.273} 
& \textbf{0.246} / \textbf{0.188} 
& 0.211 / \textbf{0.173} 
& 0.274 / 0.198 \\
\bottomrule
\end{tabular}
}
\label{tab:CLIPRenders}
\end{table}

\begin{table}[h]
\scriptsize
\centering
\caption{\textbf{Consistent image pair editing (inpainting) evaluation}.}
\resizebox{\textwidth}{!}{%
\begin{tabularx}{\textwidth}{p{0.2cm}llHcHHccc}
\toprule
& & & MEt3R $\downarrow$ & mAA, \% $\uparrow$ & TA $\uparrow$ & EC $\uparrow$ & MEt3R (object) $\downarrow$ &TA / TA dir (object) $\uparrow$ & EC (object) $\uparrow$\\
\midrule 
\multicolumn{8}{l}{\textit{DreamBooth images}\cornerarrow}\\
\midrule 
& \multirow{3}{*}{\makecell{BrushNet\\(uses masks)}}
& Independent     & 0.480 & - & 0.28   & 0.758 & 0.646        & 0.279 / 0.218         & 0.834     \\
& & Edicho          & 0.474 & - & 0.275  & 0.810 & 0.324        & \textbf{0.287 }/ 0.244         & 0.887     \\
& & \textbf{GeM-NR}   & 0.467 & - & 0.267 / 0.251  & 0.796 & 0.258 & 0.280 / \textbf{0.245} &0.899     \\
\cmidrule{2-10}
& \multirow{2}{*}{\makecell{FLUX.2\\(with masks)}}
& Independent     & 0.470 & - & 0.263 / 0.230  & 0.793 & 0.528        & 0.271  / 0.210         & 0.871     \\
& & \textbf{GeM-NR}   & 0.537 & - & 0.256 / 0.227  & 0.800 & \textbf{0.244}        & 0.268  / 0.224         & \textbf{0.902}     \\
\midrule 
\multicolumn{9}{l}{\textit{Mip-NeRF 360 test scenes}\cornerarrow}\\
\midrule
& \multirow{3}{*}{\makecell{BrushNet\\(uses masks)}}
& Independent     & 0.348 & 23.68 & 0.251 / 0.163  & 0.853 & 0.500   & \textbf{0.252 }    / 0.177 & 0.885   \\
& & Edicho          & 0.269 & 28.13 & 0.252 / 0.166  & 0.915 & 0.319 & 0.248     / 0.178 & \textbf{0.941}   \\
& & \textbf{GeM-NR}   & 0.276 & 28.04 & 0.247 / 0.165  & 0.891 & 0.285        & 0.248     / 0.179         & 0.920   \\
\cmidrule{2-10}
& \multirow{2}{*}{\makecell{FLUX.2\\(with masks)}}
& Independent     & 0.266 & 30.22 & 0.242 / 0.185  & 0.863 & 0.271        & 0.242     / 0.193         & 0.905    \\
& & \textbf{GeM-NR} & 0.228 & \textbf{33.89} & 0.236 / 0.186  & 0.892 & \textbf{0.217}        & 0.238     / \textbf{0.198}         & 0.920\\
\bottomrule
\end{tabularx}
}
\label{tab:eval_pairs}
\end{table}

\subsection{General multi-view editing}

\paragraph{Multi-view consistency.}
We evaluate multi-view consistency when editing a sequence of 10 images. Consistency metrics are reported in Table~\ref{tab:MultiViewConsistency}, both overall and per edit category. In general, our method achieves the highest consistency, with the largest improvements for the general nonrigid edits involving significant geometry changes. Omni-3DEdit is limited in scope and struggles with performing the edit for appearance change and general nonrigid edits, as seen by the text alignment scores that measure how well the edit instruction is preserved after multi-view editing. MVInpainter is limited to tasks such as object removal that can be formulated as inpainting. In contrast, our method successfully handles a broad range of edits. Figure~\ref{fig:qual_results_multi-view} shows that GeM-NR is able to successfully perform the multi-view edits, while Omni-3DEdit fails at performing the edit and independent edits lead to significant inconsistencies. The top two rows of Figure~\ref{fig:qual_results_mv_single_row} show examples where Omni-3DEdit successfully performs the edit, but GeM-NR better preserves consistent details and appearance with the anchor edit. For some edit types, e.g., appearance change, Omni-3DEdit gets better SSIM and LPIPS scores than GeM-NR. A reason for this is that these metrics can favor weaker edits that are inherently more consistent, as can be seen in our method getting higher TA scores. In Figure~\ref{fig:armchair} we also show an example of how our method can perform consistent multi-view editing for use-cases in interior design.

\paragraph{Edited 3D representations.}
The edited multi-views can also be used to generate a 3D Gaussian representing the edited scene. Table~\ref{tab:CLIPRenders} shows that GeM-NR gives edited 3D Gaussians that best align with the desired edit instruction, with the largest improvements for object addition and general nonrigid edits where there are significant changes in geometry. In Figure~\ref{fig:qual_results_renders} we observe that GeM-NR gives renders that are sharp and clearly preserve the details in the edited anchor view. The renderings from these 3D Gaussians can be found on the project page. We also show in Figure~\ref{fig:vggt} that a 3D reconstruction of the edited scene can be obtained by using VGGT \cite{vggt} on the edited images.

\subsection{Image pair editing}
Table~\ref{tab:eval_pairs} shows evaluation results for image pair editing on DreamBooth images and Mip-NeRF 360 test scenes. MEt3R, TA, and EC are computed on the masked images where only the object is kept (since the background varies a lot between the images, see top row in Figure~\ref{fig:qual_results_pairs}). GeM-NR achieves comparable or better consistency as well as text alignment.

\begin{table}[t]
\centering
\footnotesize
\caption{\textbf{Ablation study of consistent image pair editing}. Our method is highlighted in gray. It achieves a trade-off across all performance evaluation axes.}
\begin{tabular}{Hccccccccc}
\toprule
\textbf{} &
Reference text &
\multicolumn{3}{c}{Reference image(s)} &
\multicolumn{5}{c}{Metrics} \\
\cmidrule(lr){3-5}
\cmidrule(lr){6-10}
 &
\makecell{Input\\edit} &
\makecell{Original\\query} &
\makecell{Edited\\anchor} &
\makecell{Depth-\\warp} &
MEt3R $\downarrow$ &
mAA, \% $\uparrow$ &
TA $\uparrow$ &
EC $\uparrow$ &
Balanced $\uparrow$\\
\midrule
\multicolumn{9}{c}{\textbf{Simultaneous pair editing}} \\
\multicolumn{9}{l}{\textit{Concatenate both inputs}} \\
ConcatenatedPair & \checkmark & N/A & N/A & N/A & 0.229 & 35.40 & 0.271 & 0.888 & 8.08\\
\midrule
\multicolumn{9}{c}{\textbf{Editing one conditioned on another}} \\
\multicolumn{9}{l}{\textit{Original edit text prompt preserved; no warp}} \\
Independent      & \checkmark & \checkmark & \ding{55} & \ding{55}  & 0.238 & 31.70 & 0.214 & 0.924 & 7.43 \\
Baseline         & \checkmark & \checkmark & \checkmark & \ding{55} & 0.069 & 13.53 & 0.271 & 0.888 & 3.53 \\
\multicolumn{9}{l}{\textit{Original edit text prompt ignored; with warp}} \\
WarpInpaint      & \ding{55} & \ding{55}  & \ding{55}  & \checkmark & 0.219 & 39.83 & 0.268 & 0.827 & 8.47 \\
WarpInpaint2     & \ding{55} & \checkmark & \ding{55}  & \checkmark & 0.220 & 40.74 & 0.268 & 0.844 & 8.87 \\
WarpInpaint3     & \ding{55} & \ding{55}  & \checkmark & \checkmark & 0.231 & 38.74 & 0.266 & 0.823 & 8.25 \\
NoInitialPrompt  & \ding{55} & \checkmark & \checkmark & \checkmark & 0.185 & 26.11 & 0.271 & 0.863 & 5.92 \\
\multicolumn{9}{l}{\textit{Original edit text prompt preserved; with warp}} \\
WAV3D\_WithAnchor       & \checkmark & \checkmark & \checkmark & \checkmark & 0.173 & 23.96 & 0.275 & 0.913 & 5.86\\
\rowcolor{gray!30} WAV3DSingleCondPipeline & \checkmark & \checkmark & \ding{55}  & \checkmark & 0.199 & 40.63 & 0.274 & 0.885 & \textbf{9.50}\\
\multicolumn{5}{r}{\textit{No relative pose refinement}}  & 0.158 & 39.88 & 0.278 & 0.894 & 9.28     \\
\midrule
& \multicolumn{4}{r}{\textit{Unedited}} & 0.190 & 49.72 & 0.214 & 0.924 & -\\
\bottomrule
\end{tabular}
\label{tab:ablations_simple}
\end{table}

\subsection{Ablation Studies} \label{sec:ablation}
 Table~\ref{tab:ablations_simple} shows the ablation configurations and results on the held-out validation dataset. We evaluate several modifications to the proposed method: (1) Reference text (Input edit): whether to provide the original edit description; if not, provide ``Edit the first image in the same way as shown in the second image''; (2) Reference image(s): which of the images to provide: original query $Q_{src}$, edited anchor $A_{edited}$, and depth-warp $Q_{warp}$. We also compare our method to the baseline approach (first row) that concatenates the two inputs into one image and asks the model to edit the two images consistently. Finally, we evaluate GeM-NR without pose refinement (last row). GeM-NR is highlighted in gray. It achieves a trade-off across all performance evaluation metrics, as indicated in the last column that reports a balanced score: $\text{\texttt{Balanced}} = \text{\texttt{mAA}} \cdot ( 1 - \text{\texttt{MEt3R}}/2) \cdot \texttt{TA} \cdot \texttt{EC}$. The detailed instructions used for the different methods can be found in Section~B in the supplementary material.

\section{Conclusion}
\label{sec:conclusion}
We present GeM-NR, a method for multi-view consistent edits that can 
substantially change scene geometry and appearance. Our approach is flexible, handling diverse editing tasks, and fast, requiring only a few seconds per image. Extensive evaluations on different edit categories show that GeM-NR improves multi-view consistency and that the edited images can be used to generate 3D representations that align well with the specified edit instruction. 

Our current approach conditions multi-view editing on a single anchor edit, which does not explicitly enforce consistency between query views. While this strategy works well for scenes with limited viewpoint variation, it becomes less effective for larger scenes with more extreme viewpoint changes. A potential solution is to condition the editing not only on one anchor view, but also on other previously edited images.

\newpage
\subsection*{Acknowledgments}
This work was supported by the Wallenberg AI, Autonomous Systems, and Software Program (WASP) funded by the Knut and Alice Wallenberg Foundation. Computational resources were provided by the National Academic Infrastructure for Supercomputing in Sweden (NAISS) at Chalmers Centre for Computational Science and Engineering (C3SE), partially funded by the Swedish Research Council under grant agreement no. 2022-06725, and by the Berzelius resource, provided by the Knut and Alice Wallenberg Foundation at the National Supercomputer
Centre.

\bibliographystyle{splncs04}
\bibliography{egbib}

@inproceedings{diffusion,
author = {Ho, Jonathan and Jain, Ajay and Abbeel, Pieter},
title = {Denoising diffusion probabilistic models},
year = {2020},
isbn = {9781713829546},
publisher = {Curran Associates Inc.},
address = {Red Hook, NY, USA},
booktitle = {Proceedings of the 34th International Conference on Neural Information Processing Systems},
articleno = {574},
numpages = {12},
location = {Vancouver, BC, Canada},
series = {NIPS '20}
}

@InProceedings{stable_diffusion,
    author    = {Rombach, Robin and Blattmann, Andreas and Lorenz, Dominik and Esser, Patrick and Ommer, Bj\"orn},
    title     = {High-Resolution Image Synthesis With Latent Diffusion Models},
    booktitle = {Proceedings of the IEEE/CVF Conference on Computer Vision and Pattern Recognition (CVPR)},
    month     = {June},
    year      = {2022},
    pages     = {10684-10695}
}

@article{barron2022mipnerf360,
    title={Mip-NeRF 360: Unbounded Anti-Aliased Neural Radiance Fields},
    author={Jonathan T. Barron and Ben Mildenhall and 
            Dor Verbin and Pratul P. Srinivasan and Peter Hedman},
    journal={Proceedings of the Computer Vision and Pattern Recognition Conference},
    year={2022}
}

@article{blendedmvs,
  author       = {Yao Yao and
                  Zixin Luo and
                  Shiwei Li and
                  Jingyang Zhang and
                  Yufan Ren and
                  Lei Zhou and
                  Tian Fang and
                  Long Quan},
  title        = {BlendedMVS: {A} Large-scale Dataset for Generalized Multi-view Stereo
                  Networks},
  journal={Proceedings of the Computer Vision and Pattern Recognition Conference},
  year={2020}
}

@inproceedings{imagen,
 author = {Saharia, Chitwan and Chan, William and Saxena, Saurabh and Li, Lala and Whang, Jay and Denton, Emily L and Ghasemipour, Kamyar and Gontijo Lopes, Raphael and Karagol Ayan, Burcu and Salimans, Tim and Ho, Jonathan and Fleet, David and Norouzi, Mohammad},
 booktitle = {Advances in Neural Information Processing Systems},
 editor = {S. Koyejo and S. Mohamed and A. Agarwal and D. Belgrave and K. Cho and A. Oh},
 pages = {36479--36494},
 publisher = {Curran Associates, Inc.},
 title = {Photorealistic Text-to-Image Diffusion Models with Deep Language Understanding},
 url = {https://proceedings.neurips.cc/paper_files/paper/2022/file/ec795aeadae0b7d230fa35cbaf04c041-Paper-Conference.pdf},
 volume = {35},
 year = {2022}
}

@misc{dalle2,
      title={Hierarchical Text-Conditional Image Generation with CLIP Latents}, 
      author={Aditya Ramesh and Prafulla Dhariwal and Alex Nichol and Casey Chu and Mark Chen},
      year={2022},
      eprint={2204.06125},
      archivePrefix={arXiv},
      primaryClass={cs.CV},
      url={https://arxiv.org/abs/2204.06125}, 
}

@InProceedings{instructpix2pix,
    author     = {Brooks, Tim and Holynski, Aleksander and Efros, Alexei A.},
    title      = {InstructPix2Pix: Learning to Follow Image Editing Instructions},
    booktitle  = {Proceedings of the Computer Vision and Pattern Recognition Conference},
    year       = {2023},
}

@inproceedings{sdedit,
  title     = {SDEdit: Guided Image Synthesis and Editing with Stochastic Differential Equations},
  author    = {Meng, Chenlin and He, Yutong and Song, Yang and Song, Jiaming and Wu, Jiajun and Zhu, Jun-Yan and Ermon, Stefano},
  booktitle = {International Conference on Learning Representations},
  year      = {2022}
}

@InProceedings{dream_booth,
    author    = {Ruiz, Nataniel and Li, Yuanzhen and Jampani, Varun and Pritch, Yael and Rubinstein, Michael and Aberman, Kfir},
    title     = {DreamBooth: Fine Tuning Text-to-Image Diffusion Models for Subject-Driven Generation},
    booktitle = {Proceedings of the IEEE/CVF Conference on Computer Vision and Pattern Recognition (CVPR)},
    month     = {June},
    year      = {2023},
    pages     = {22500-22510}
}

@inproceedings{prompt2prompt,
  title     = {Prompt-to-Prompt Image Editing with Cross-Attention Control},
  author    = {Hertz, Amir and Mokady, Ron and Tenenbaum, Jay and Aberman, Kfir and Pritch, Yael and Cohen-Or, Daniel},
  booktitle = {International Conference on Learning Representations},
  year      = {2023}
}

@inproceedings{magic_brush,
 author = {Zhang, Kai and Mo, Lingbo and Chen, Wenhu and Sun, Huan and Su, Yu},
 booktitle = {Advances in Neural Information Processing Systems},
 editor = {A. Oh and T. Naumann and A. Globerson and K. Saenko and M. Hardt and S. Levine},
 pages = {31428--31449},
 publisher = {Curran Associates, Inc.},
 title = {MagicBrush: A Manually Annotated Dataset for Instruction-Guided Image Editing},
 url = {https://proceedings.neurips.cc/paper_files/paper/2023/file/64008fa30cba9b4d1ab1bd3bd3d57d61-Paper-Datasets_and_Benchmarks.pdf},
 volume = {36},
 year = {2023}
}

@inproceedings{mgie,
title = {Guiding Instruction-based Image Editing via Multimodal Large Language Models},
booktitle = {ICLR},
author = {Tsu-Jui Fu and Wenze Hu and Xianzhi Du and William Wang and Yinfei Yang and Zhe Gan},
year = {2024},
URL = {https://arxiv.org/abs/2309.17102}
}

@InProceedings{repaint,
    author    = {Lugmayr, Andreas and Danelljan, Martin and Romero, Andres and Yu, Fisher and Timofte, Radu and Van Gool, Luc},
    title     = {RePaint: Inpainting Using Denoising Diffusion Probabilistic Models},
    booktitle = {Proceedings of the IEEE/CVF Conference on Computer Vision and Pattern Recognition (CVPR)},
    month     = {June},
    year      = {2022},
    pages     = {11461-11471}
}

@InProceedings{brushnet,
author="Ju, Xuan
and Liu, Xian
and Wang, Xintao
and Bian, Yuxuan
and Shan, Ying
and Xu, Qiang",
editor="Leonardis, Ale{\v{s}}
and Ricci, Elisa
and Roth, Stefan
and Russakovsky, Olga
and Sattler, Torsten
and Varol, G{\"u}l",
title="BrushNet: A Plug-and-Play Image Inpainting Model with Decomposed Dual-Branch Diffusion",
booktitle="Computer Vision -- ECCV 2024",
year="2025",
publisher="Springer Nature Switzerland",
address="Cham",
pages="150--168",
}

@InProceedings{PowerPaint,
author="Zhuang, Junhao
and Zeng, Yanhong
and Liu, Wenran
and Yuan, Chun
and Chen, Kai",
editor="Leonardis, Ale{\v{s}}
and Ricci, Elisa
and Roth, Stefan
and Russakovsky, Olga
and Sattler, Torsten
and Varol, G{\"u}l",
title="A Task Is Worth One Word: Learning with Task Prompts for High-Quality Versatile Image Inpainting",
booktitle="Computer Vision -- ECCV 2024",
year="2025",
publisher="Springer Nature Switzerland",
address="Cham",
pages="195--211",
}

@InProceedings{inst,
 author    = {Zhang, Yuxin and Huang, Nisha and Tang, Fan and Huang, Haibin and Ma, Chongyang and Dong, Weiming and Xu, Changsheng},
 title     = {Inversion-Based Style Transfer With Diffusion Models},
 booktitle = {Proceedings of the IEEE/CVF Conference on Computer Vision and Pattern Recognition (CVPR)},
 month     = {June},
 year      = {2023},
 pages     = {10146-10156}
}

@InProceedings{style_aligned,
    author    = {Hertz, Amir and Voynov, Andrey and Fruchter, Shlomi and Cohen-Or, Daniel},
    title     = {Style Aligned Image Generation via Shared Attention},
    booktitle = {Proceedings of the IEEE/CVF Conference on Computer Vision and Pattern Recognition (CVPR)},
    month     = {June},
    year      = {2024},
    pages     = {4775-4785}
}

@InProceedings{style_injection,
    author    = {Chung, Jiwoo and Hyun, Sangeek and Heo, Jae-Pil},
    title     = {Style Injection in Diffusion: A Training-free Approach for Adapting Large-scale Diffusion Models for Style Transfer},
    booktitle = {Proceedings of the IEEE/CVF Conference on Computer Vision and Pattern Recognition (CVPR)},
    month     = {June},
    year      = {2024},
    pages     = {8795-8805}
}

@InProceedings{art_adapter,
    author    = {Chen, Dar-Yen and Tennent, Hamish and Hsu, Ching-Wen},
    title     = {ArtAdapter: Text-to-Image Style Transfer using Multi-Level Style Encoder and Explicit Adaptation},
    booktitle = {Proceedings of the IEEE/CVF Conference on Computer Vision and Pattern Recognition (CVPR)},
    month     = {June},
    year      = {2024},
    pages     = {8619-8628}
}

@inproceedings{
flow-matching,
title={Flow Matching for Generative Modeling},
author={Yaron Lipman and Ricky T. Q. Chen and Heli Ben-Hamu and Maximilian Nickel and Matthew Le},
booktitle={The Eleventh International Conference on Learning Representations },
year={2023},
url={https://openreview.net/forum?id=PqvMRDCJT9t}
}

@inproceedings{
rectified_flow,
title={Flow Straight and Fast: Learning to Generate and Transfer Data with Rectified Flow},
author={Xingchao Liu and Chengyue Gong and qiang liu},
booktitle={The Eleventh International Conference on Learning Representations },
year={2023},
url={https://openreview.net/forum?id=XVjTT1nw5z}
}

@InProceedings{mmdit,
  title = 	 {Scaling Rectified Flow Transformers for High-Resolution Image Synthesis},
  author =       {Esser, Patrick and Kulal, Sumith and Blattmann, Andreas and Entezari, Rahim and M\"{u}ller, Jonas and Saini, Harry and Levi, Yam and Lorenz, Dominik and Sauer, Axel and Boesel, Frederic and Podell, Dustin and Dockhorn, Tim and English, Zion and Rombach, Robin},
  booktitle = 	 {Proceedings of the 41st International Conference on Machine Learning},
  pages = 	 {12606--12633},
  year = 	 {2024},
  volume = 	 {235},
  series = 	 {Proceedings of Machine Learning Research},
  month = 	 {21--27 Jul},
  publisher =    {PMLR},
  url = 	 {https://proceedings.mlr.press/v235/esser24a.html},
}

@misc{flux1,
      title={FLUX.1 Kontext: Flow Matching for In-Context Image Generation and Editing in Latent Space},
      author={Black Forest Labs and Stephen Batifol and Andreas Blattmann and Frederic Boesel and Saksham Consul and Cyril Diagne and Tim Dockhorn and Jack English and Zion English and Patrick Esser and Sumith Kulal and Kyle Lacey and Yam Levi and Cheng Li and Dominik Lorenz and Jonas Müller and Dustin Podell and Robin Rombach and Harry Saini and Axel Sauer and Luke Smith},
      year={2025},
      eprint={2506.15742},
      archivePrefix={arXiv},
      primaryClass={cs.GR},
      url={https://arxiv.org/abs/2506.15742},
}

@misc{flux2,
  author       = {{Black Forest Labs}},
  title        = {{FLUX.2: Frontier Visual Intelligence}},
  year         = {2025},
  howpublished = {\url{https://bfl.ai/blog/flux-2}}
}

@misc{flux2klein,
  author       = {{Black Forest Labs}},
  title        = {{FLUX.2 [klein]: Towards Interactive Visual Intelligence}},
  year         = {2025},
  howpublished = {\url{https://bfl.ai/blog/flux2-klein-towards-interactive-visual-intelligence}}
}

@misc{qwen,
      title={Qwen-Image Technical Report}, 
      author={Chenfei Wu and Jiahao Li and Jingren Zhou and Junyang Lin and Kaiyuan Gao and Kun Yan and Sheng-ming Yin and Shuai Bai and Xiao Xu and Yilei Chen and Yuxiang Chen and Zecheng Tang and Zekai Zhang and Zhengyi Wang and An Yang and Bowen Yu and Chen Cheng and Dayiheng Liu and Deqing Li and Hang Zhang and Hao Meng and Hu Wei and Jingyuan Ni and Kai Chen and Kuan Cao and Liang Peng and Lin Qu and Minggang Wu and Peng Wang and Shuting Yu and Tingkun Wen and Wensen Feng and Xiaoxiao Xu and Yi Wang and Yichang Zhang and Yongqiang Zhu and Yujia Wu and Yuxuan Cai and Zenan Liu},
      year={2025},
      eprint={2508.02324},
      archivePrefix={arXiv},
      primaryClass={cs.CV},
      url={https://arxiv.org/abs/2508.02324}, 
}

@InProceedings{omnigen,
    author    = {Xiao, Shitao and Wang, Yueze and Zhou, Junjie and Yuan, Huaying and Xing, Xingrun and Yan, Ruiran and Li, Chaofan and Wang, Shuting and Huang, Tiejun and Liu, Zheng},
    title     = {OmniGen: Unified Image Generation},
    booktitle = {Proceedings of the IEEE/CVF Conference on Computer Vision and Pattern Recognition (CVPR)},
    month     = {June},
    year      = {2025},
    pages     = {13294-13304}
}

@misc{chen2023gaussianeditor,
    title={GaussianEditor: Swift and Controllable 3D Editing with Gaussian Splatting},
    author={Yiwen Chen and Zilong Chen and Chi Zhang and Feng Wang and Xiaofeng Yang and Yikai Wang and Zhongang Cai and Lei Yang and Huaping Liu and Guosheng Lin},
    year={2024},
booktitle={Proceedings of the Computer Vision and Pattern Recognition Conference},
}

@inproceedings{mirzaei2023watchyoursteps,
  title={Watch Your Steps: Local Image and Scene Editing by Text Instructions}, 
  author={Ashkan Mirzaei and Tristan Aumentado-Armstrong and Marcus A. Brubaker and Jonathan Kelly and Alex Levinshtein and Konstantinos G. Derpanis and Igor Gilitschenski},
  year={2024},
  booktitle={ECCV},
}

@article{ProteusNeRF,
author = {Wang, Binglun and Dutt, Niladri Shekhar and Mitra, Niloy J.},
title = {ProteusNeRF: Fast Lightweight NeRF Editing using 3D-Aware Image Context},
year = {2024},
issue_date = {May 2024},
publisher = {Association for Computing Machinery},
address = {New York, NY, USA},
volume = {7},
number = {1},
url = {https://doi.org/10.1145/3651290},
doi = {10.1145/3651290},
journal = {Proc. ACM Comput. Graph. Interact. Tech.},
month = {may},
articleno = {22},
numpages = {17}
}

@inproceedings{vcedit,
author = {Wang, Yuxuan and Yi, Xuanyu and Wu, Zike and Zhao, Na and Chen, Long and Zhang, Hanwang},
title = {View-Consistent 3D Editing with Gaussian Splatting},
year = {2024},
isbn = {978-3-031-72760-3},
publisher = {Springer-Verlag},
address = {Berlin, Heidelberg},
url = {https://doi.org/10.1007/978-3-031-72761-0_23},
doi = {10.1007/978-3-031-72761-0_23},
booktitle = {Computer Vision – ECCV 2024: 18th European Conference, Milan, Italy, September 29 – October 4, 2024, Proceedings,  Part XXXV},
pages = {404–420},
numpages = {17},
location = {Milan, Italy}
}

@inproceedings{haque2023instruct,
  title={Instruct-nerf2nerf: Editing 3d scenes with instructions},
  author={Haque, Ayaan and Tancik, Matthew and Efros, Alexei A and Holynski, Aleksander and Kanazawa, Angjoo},
  booktitle={Proceedings of the IEEE/CVF international conference on computer vision},
  pages={19740--19750},
  year={2023}
}

@inproceedings{editsplat,
  title={Editsplat: Multi-view fusion and attention-guided optimization for view-consistent 3d scene editing with 3d gaussian splatting},
  author={Lee, Dong In and Park, Hyeongcheol and Seo, Jiyoung and Park, Eunbyung and Park, Hyunje and Baek, Ha Dam and Shin, Sangheon and Kim, Sangmin and Kim, Sangpil},
  booktitle={Proceedings of the Computer Vision and Pattern Recognition Conference},
  pages={11135--11145},
  year={2025}
}

@inproceedings{dge,
  title={Dge: Direct gaussian 3d editing by consistent multi-view editing},
  author={Chen, Minghao and Laina, Iro and Vedaldi, Andrea},
  booktitle={European Conference on Computer Vision},
  pages={74--92},
  year={2024},
  organization={Springer}
}

@article{gaussctrl,
author = {Wu, Jing and Bian, Jia-Wang and Li, Xinghui and Wang, Guangrun and Reid, Ian and Torr, Philip and Prisacariu, Victor},
title = {{GaussCtrl: Multi-View Consistent Text-Driven 3D Gaussian Splatting Editing}},
journal = {ECCV},
year = {2024},
}

@ARTICLE{core_editor,
  author={Zhu, Zhe and Chen, Honghua and Li, Peng and Wei, Mingqiang},
  journal={IEEE Transactions on Visualization and Computer Graphics}, 
  title={CoreEditor: Correspondence-Constrained Diffusion for Consistent 3D Editing}, 
  year={2026},
  volume={32},
  number={3},
  pages={2838-2851},
  doi={10.1109/TVCG.2026.3657658}}

@article{vip3de, title={Fast Multi-view Consistent 3D Editing with Video Priors}, volume={40}, url={https://ojs.aaai.org/index.php/AAAI/article/view/37286}, DOI={10.1609/aaai.v40i4.37286}, abstractNote={Text-driven 3D editing enables user-friendly 3D object or scene editing with text instructions. Due to the lack of multi-view consistency priors, existing methods typically resort to employ 2D generation or editing models to process per-view individually, followed by iterative 2D-3D-2D updating. However, these methods are not only time-consuming but also prone to yielding over-smoothed results, since iterative process averages the different editing signals gathered from different views. In this paper, we propose, an early and pioneering work of generative Video Prior based 3D Editing, ViP3DE in short, to repurpose the temporal consistency priors from pre-trained video generation models to achieve consistent 3D editing within a single forward pass. Our key insight is to condition the video generation model on a single edited view to generate other consistent edited views for 3D updating directly, thereby bypassing iterative editing paradigm. First, 3D updating requires edited views to be paired with specific camera poses. To this end, we propose \textit{motion-preserved noise blending} for the video model to generate edited views at predefined camera poses. In addition, we introduce \textit{geometrically aware denoising} to further enhance multi-view consistency by integrating 3D geometric priors into video models. Extensive experiments demonstrate that our proposed ViP3DE can achieve high-quality 3D editing results even within a single forward pass, significantly outperforming existing methods in both editing quality and editing time cost.}, number={4}, journal={Proceedings of the AAAI Conference on Artificial Intelligence}, author={Chen, Liyi and Li, Ruihuang and Zhang, Guowen and Wang, Pengfei and Zhang, Lei}, year={2026}, month={Mar.}, pages={2948–2956} }

@InProceedings{edicho,
    author    = {Bai, Qingyan and Ouyang, Hao and Xu, Yinghao and Wang, Qiuyu and Yang, Ceyuan and Cheng, Ka Leong and Shen, Yujun and Chen, Qifeng},
    title     = {Edicho: Consistent Image Editing in the Wild},
    booktitle = {Proceedings of the IEEE/CVF International Conference on Computer Vision (ICCV)},
    month     = {October},
    year      = {2025},
    pages     = {15277-15287}
}

@article{efficient-nerf2nerf,
    title={Efficient-NeRF2NeRF: Streamlining Text-Driven 3D Editing 
           with Multiview Correspondence-Enhanced Diffusion Models}, 
    author={Liangchen Song and Liangliang Cao and Jiatao Gu and Yifan Jiang and Junsong Yuan and Hao Tang},
    journal={arXiv preprint arXiv:2312.08563},
    year={2023}
}

@inproceedings{correspondence_guidance,
      title={3D-Consistent Multi-View Editing by Correspondence Guidance}, 
      author={Josef Bengtson and David Nilsson and Dong In Lee and Yaroslava Lochman and Fredrik Kahl},
      booktitle={European Conference on Computer Vision (ECCV) Workshops},
      year={2026},
}

@misc{ye2025nano3dtrainingfreeapproachefficient,
      title={NANO3D: A Training-Free Approach for Efficient 3D Editing Without Masks}, 
      author={Junliang Ye and Shenghao Xie and Ruowen Zhao and Zhengyi Wang and Hongyu Yan and Wenqiang Zu and Lei Ma and Jun Zhu},
      year={2025},
      eprint={2510.15019},
      archivePrefix={arXiv},
      primaryClass={cs.CV},
      url={https://arxiv.org/abs/2510.15019}, 
}

@misc{xia2025scalableconsistent3dediting,
      title={Towards Scalable and Consistent 3D Editing}, 
      author={Ruihao Xia and Yang Tang and Pan Zhou},
      year={2025},
      eprint={2510.02994},
      archivePrefix={arXiv},
      primaryClass={cs.CV},
      url={https://arxiv.org/abs/2510.02994}, 
}

@InProceedings{ava-nvs,
author="Bengtson, Josef
and Nilsson, David
and Lin, Che-Tsung
and B{\"u}sching, Marcel
and Kahl, Fredrik",
editor="Wallraven, Christian
and Liu, Cheng-Lin
and Ross, Arun",
title="Adjustable Visual Appearance for Generalizable Novel View Synthesis",
booktitle="Pattern Recognition and Artificial Intelligence",
year="2025",
publisher="Springer Nature Singapore",
address="Singapore",
pages="157--171",
isbn="978-981-97-8702-9"
}

@inproceedings{omni3dedit,
  title     = {Omni-3DEdit: Generalized Versatile 3D Editing in One-Pass},
  author={Liyi, Chen and Pengfei, Wang and Guowen, Zhang and Zhiyuan, Ma and Lei, Zhang},
  booktitle = {Proceedings of the IEEE/CVF Conference on Computer Vision and Pattern Recognition (CVPR)},
  year      = {2026},
}

@article{li2025voxhammer,
    title = {VoxHammer: Training-Free Precise and Coherent 3D Editing in Native 3D Space},
    author = {Li, Lin and Huang, Zehuan and Feng, Haoran and Zhuang, Gengxiong and Chen, Rui and Guo, Chunchao and Sheng, Lu},
    journal = {arXiv preprint arXiv:2508.19247},
    year = {2025}
}

@article{rl3dedit,
  title={Geometry-Guided Reinforcement Learning for Multi-view Consistent 3D Scene Editing},
  author={Wang, Jiyuan and Lin, Chunyu and Sun, Lei and Cao, Zhi and Yin, Yuyang and Nie, Lang and Yuan, Zhenlong and Chu, Xiangxiang and Wei, Yunchao and Liao, Kang and others},
  journal={arXiv preprint arXiv:2603.03143},
  year={2026}
}

@inproceedings{vggt,
  title={VGGT: Visual Geometry Grounded Transformer},
  author={Wang, Jianyuan and Chen, Minghao and Karaev, Nikita and Vedaldi, Andrea and Rupprecht, Christian and Novotny, David},
  booktitle={Proceedings of the IEEE/CVF Conference on Computer Vision and Pattern Recognition},
  year={2025}
}

@inproceedings{
  megascenes,
  title={MegaScenes: Scene-Level View Synthesis at Scale}, 
  author={Tung, Joseph and Chou, Gene and Cai, Ruojin and Yang, Guandao and Zhang, Kai and Wetzstein, Gordon and Hariharan, Bharath and Snavely, Noah},
  booktitle={ECCV},
  year={2024}
}

@InProceedings{infinite_nature,
  author = {Liu, Andrew and Tucker, Richard and Jampani, Varun and
            Makadia, Ameesh and Snavely, Noah and Kanazawa, Angjoo},
  title = {Infinite Nature: Perpetual View Generation of Natural Scenes from a Single Image},
  booktitle = {Proceedings of the IEEE/CVF International Conference on Computer Vision (ICCV)},
  month = {October},
  year = {2021}
}

@ARTICLE{viewcrafter,
author={Yu, Wangbo and Xing, Jinbo and Yuan, Li and Hu, Wenbo and Li, Xiaoyu and Huang, Zhipeng and Gao, Xiangjun and Wong, Tien-Tsin and Shan, Ying and Tian, Yonghong},
journal={ IEEE Transactions on Pattern Analysis \& Machine Intelligence },
title={{ ViewCrafter: Taming Video Diffusion Models for High-fidelity Novel View Synthesis }},
year={2025},
volume={},
number={01},
ISSN={1939-3539},
pages={1-18},
doi={10.1109/TPAMI.2025.3613256},
url = {https://doi.ieeecomputersociety.org/10.1109/TPAMI.2025.3613256},
publisher={IEEE Computer Society},
address={Los Alamitos, CA, USA},
month=sep}

@InProceedings{multidiff,
                author    = {M\"uller, Norman and Schwarz, Katja and R\"ossle, Barbara and Porzi, Lorenzo and Bul\`o, Samuel Rota and Nie{\ss}ner, Matthias and Kontschieder, Peter},
                title     = {MultiDiff: Consistent Novel View Synthesis from a Single Image},
                booktitle = {Proceedings of the IEEE/CVF Conference on Computer Vision and Pattern Recognition (CVPR)},
                month     = {June},
                year      = {2024},
                pages     = {10258-10268}
            }

@inproceedings{vica_nerf,
  title={ViCA-NeRF: View-Consistency-Aware 3D Editing of Neural Radiance Fields},
  author={Dong, Jiahua and Wang, Yu-Xiong},
  booktitle={Thirty-seventh Conference on Neural Information Processing Systems},
  year={2023}
}

@inproceedings{date_nerf,
author = {Rojas, Sara and Philip, Julien and Zhang, Kai and Bi, Sai and Luan, Fujun and Ghanem, Bernard and Sunkavalli, Kalyan},
title = {DATENeRF: Depth-Aware Text-Based Editing of NeRFs},
year = {2024},
isbn = {978-3-031-73246-1},
publisher = {Springer-Verlag},
address = {Berlin, Heidelberg},
url = {https://doi.org/10.1007/978-3-031-73247-8_16},
doi = {10.1007/978-3-031-73247-8_16},
booktitle = {Computer Vision – ECCV 2024: 18th European Conference, Milan, Italy, September 29–October 4, 2024, Proceedings, Part XI},
pages = {267–284},
numpages = {18},
location = {Milan, Italy}
}

@inproceedings{
DA3,
title={Depth Anything 3: Recovering the Visual Space from Any Views},
author={Haotong Lin and Sili Chen and Jun Hao Liew and Donny Y. Chen and Zhenyu Li and Yang Zhao and Sida Peng and Hengkai Guo and Xiaowei Zhou and Guang Shi and Jiashi Feng and Bingyi Kang},
booktitle={The Fourteenth International Conference on Learning Representations},
year={2026},
url={https://openreview.net/forum?id=yirunib8l8}
}

@ARTICLE {midas,
    author  = "Ren\'{e} Ranftl and Katrin Lasinger and David Hafner and Konrad Schindler and Vladlen Koltun",
    title   = "Towards Robust Monocular Depth Estimation: Mixing Datasets for Zero-Shot Cross-Dataset Transfer",
    journal = "IEEE Transactions on Pattern Analysis and Machine Intelligence",
    year    = "2022",
    volume  = "44",
    number  = "3"
}

@InProceedings{dpt,
    author    = {Ranftl, Ren\'e and Bochkovskiy, Alexey and Koltun, Vladlen},
    title     = {Vision Transformers for Dense Prediction},
    booktitle = {Proceedings of the IEEE/CVF International Conference on Computer Vision (ICCV)},
    month     = {October},
    year      = {2021},
    pages     = {12179-12188}
}

@inproceedings{DA,
  title={Depth Anything: Unleashing the Power of Large-Scale Unlabeled Data},
  author={Yang, Lihe and Kang, Bingyi and Huang, Zilong and Xu, Xiaogang and Feng, Jiashi and Zhao, Hengshuang},
  booktitle={CVPR},
  year={2024}
}

@misc{gemini2_5,
      title={Gemini 2.5: Pushing the Frontier with Advanced Reasoning, Multimodality, Long Context, and Next Generation Agentic Capabilities}, 
      author={Gheorghe Comanici and Eric Bieber and Mike Schaekermann and Ice Pasupat and Noveen Sachdeva and Inderjit Dhillon and Marcel Blistein and Ori Ram and Dan Zhang and Evan Rosen et. al.},
      year={2025},
      eprint={2507.06261},
      archivePrefix={arXiv},
      primaryClass={cs.CL},
      url={https://arxiv.org/abs/2507.06261}, 
}

@misc{qwen2,
      title={Qwen2 Technical Report}, 
      author={An Yang and Baosong Yang and Binyuan Hui and Bo Zheng and Bowen Yu and Chang Zhou and Chengpeng Li and Chengyuan Li and Dayiheng Liu and Fei Huang et. al.},
      year={2024},
      eprint={2407.10671},
      archivePrefix={arXiv},
      primaryClass={cs.CL},
      url={https://arxiv.org/abs/2407.10671}, 
}

@inproceedings{roma,
  title={Roma: Robust dense feature matching},
  author={Edstedt, Johan and Sun, Qiyu and B{\"o}kman, Georg and Wadenb{\"a}ck, M{\aa}rten and Felsberg, Michael},
  booktitle={Proceedings of the IEEE/CVF Conference on Computer Vision and Pattern Recognition},
  pages={19790--19800},
  year={2024}
}

@inproceedings{gen_warp,
 author = {Seo, Junyoung and Fukuda, Kazumi and Shibuya, Takashi and Narihira, Takuya and Murata, Naoki and Hu, Shoukang and Lai, Chieh-Hsin and Kim, Seungryong and Mitsufuji, Yuki},
 booktitle = {Advances in Neural Information Processing Systems},
 doi = {10.52202/079017-2550},
 editor = {A. Globerson and L. Mackey and D. Belgrave and A. Fan and U. Paquet and J. Tomczak and C. Zhang},
 pages = {80220--80243},
 publisher = {Curran Associates, Inc.},
 title = {GenWarp: Single Image to Novel Views with Semantic-Preserving Generative Warping},
 url = {https://proceedings.neurips.cc/paper_files/paper/2024/file/92e886487a8354b03d8bf4416eae6d7d-Paper-Conference.pdf},
 volume = {37},
 year = {2024}
}

@inproceedings{nvs-solver,
title={NVS-Solver: Video Diffusion Model as Zero-Shot Novel View Synthesizer},
author={You, Meng and Zhu, Zhiyu and Liu, Hui and Hou, Junhui},
booktitle={International Conference on Learning Representations},
year={2025}
}

@InProceedings{wave,
    author    = {Park, Jiwoo and Choi, Tae Eun and Jun, Youngjun and Hwang, Seong Jae},
    title     = {WAVE: Warp-Based View Guidance for Consistent Novel View Synthesis Using a Single Image},
    booktitle = {Proceedings of the IEEE/CVF International Conference on Computer Vision (ICCV)},
    month     = {October},
    year      = {2025},
    pages     = {11906-11915}
}

@misc{attention-warp,
title={Diffusion-Based Attention Warping for Consistent 3D Scene Editing}, 
author={Eyal Gomel and Lior Wolf},
year={2024},
eprint={2412.07984},
archivePrefix={arXiv},
primaryClass={cs.CV},
url={https://arxiv.org/abs/2412.07984}, 
}

@InProceedings{clip,
  title = 	 {Learning Transferable Visual Models From Natural Language Supervision},
  author =       {Radford, Alec and Kim, Jong Wook and Hallacy, Chris and Ramesh, Aditya and Goh, Gabriel and Agarwal, Sandhini and Sastry, Girish and Askell, Amanda and Mishkin, Pamela and Clark, Jack and Krueger, Gretchen and Sutskever, Ilya},
  booktitle = 	 {Proceedings of the 38th International Conference on Machine Learning},
  pages = 	 {8748--8763},
  year = 	 {2021},
  editor = 	 {Meila, Marina and Zhang, Tong},
  volume = 	 {139},
  series = 	 {Proceedings of Machine Learning Research},
  month = 	 {18--24 Jul},
  publisher =    {PMLR},
  url = 	 {https://proceedings.mlr.press/v139/radford21a.html}
}

@article{anysplat,
  title={Anysplat: Feed-forward 3d gaussian splatting from unconstrained views},
  author={Jiang, Lihan and Mao, Yucheng and Xu, Linning and Lu, Tao and Ren, Kerui and Jin, Yichen and Xu, Xudong and Yu, Mulin and Pang, Jiangmiao and Zhao, Feng and others},
  journal={ACM Transactions on Graphics (TOG)},
  volume={44},
  number={6},
  pages={1--16},
  year={2025},
  publisher={ACM New York, NY, USA}
}

@inproceedings{met3r,
  title={Met3r: Measuring multi-view consistency in generated images},
  author={Asim, Mohammad and Wewer, Christopher and Wimmer, Thomas and Schiele, Bernt and Lenssen, Jan Eric},
  booktitle={Proceedings of the Computer Vision and Pattern Recognition Conference},
  pages={6034--6044},
  year={2025}
}

@article{dino,
  title={Emerging Properties in Self-Supervised Vision Transformers},
  author={Mathilde Caron and Hugo Touvron and Ishan Misra and Herv'e J'egou and Julien Mairal and Piotr Bojanowski and Armand Joulin},
  journal={2021 IEEE/CVF International Conference on Computer Vision (ICCV)},
  year={2021},
  pages={9630-9640},
  url={https://api.semanticscholar.org/CorpusID:233444273}
}

@inproceedings{dust3r,
  title={Dust3r: Geometric 3d vision made easy},
  author={Wang, Shuzhe and Leroy, Vincent and Cabon, Yohann and Chidlovskii, Boris and Revaud, Jerome},
  booktitle={Proceedings of the IEEE/CVF Conference on Computer Vision and Pattern Recognition},
  pages={20697--20709},
  year={2024}
}

@inproceedings{spinnerf,
      title={{SPIn-NeRF}: Multiview Segmentation and Perceptual Inpainting with Neural Radiance Fields}, 
      author={Ashkan Mirzaei and Tristan Aumentado-Armstrong and Konstantinos G. Derpanis and Jonathan Kelly and Marcus A. Brubaker and Igor Gilitschenski and Alex Levinshtein},
      year={2023},
      booktitle={CVPR},
}

@misc{controlnet,
  title={Adding Conditional Control to Text-to-Image Diffusion Models}, 
  author={Lvmin Zhang and Anyi Rao and Maneesh Agrawala},
  booktitle={IEEE International Conference on Computer Vision (ICCV)},
  year={2023},
}

@article{uni-controlnet,
  title={Uni-ControlNet: All-in-One Control to Text-to-Image Diffusion Models},
  author={Zhao, Shihao and Chen, Dongdong and Chen, Yen-Chun and Bao, Jianmin and Hao, Shaozhe and Yuan, Lu and Wong, Kwan-Yee~K.},
  journal={Advances in Neural Information Processing Systems},
  year={2023}
}

@inproceedings{
diffusion_feature_field,
title={Diffusion Feature Field for Text-based 3D Editing with Gaussian Splatting},
author={Eunseo Koh and Sangeek Hyun and MinKyu Lee and Jiwoo Chung and Kangmin Seo and Jae-Pil Heo},
booktitle={The Thirty-ninth Annual Conference on Neural Information Processing Systems},
year={2025},
url={https://openreview.net/forum?id=Kf9eNbp4wy}
}

@InProceedings{Dynamic-eDiTor,
  author    = {Lee, Dong In and Doh, Hyungjun and Chi, Seunggeun and Duan, Runlin and Kim, Sangpil and Ramani, Karthik},
  title     = {Dynamic-eDiTor: Training-Free Text-Driven 4D Scene Editing with Multimodal Diffusion Transformer},
  booktitle = {Proceedings of the IEEE/CVF Conference on Computer Vision and Pattern Recognition (CVPR)},
  year      = {2026}
}

@misc{qu2025editcast3dsingleframeguided3dediting,
      title={EditCast3D: Single-Frame-Guided 3D Editing with Video Propagation and View Selection}, 
      author={Huaizhi Qu and Ruichen Zhang and Shuqing Luo and Luchao Qi and Zhihao Zhang and Xiaoming Liu and Roni Sengupta and Tianlong Chen},
      year={2025},
      eprint={2510.13652},
      archivePrefix={arXiv},
      primaryClass={cs.CV},
      url={https://arxiv.org/abs/2510.13652}, 
}

@inproceedings{mvinpainter,
 author = {Cao, Chenjie and others},
 title = {MVInpainter: Learning Multi-View Consistent Inpainting to Bridge 2D and 3D Editing},
 booktitle = {NeurIPS},
 year = {2024}
}

\setcounter{equation}{0}
\setcounter{table}{0}
\setcounter{figure}{0}
\renewcommand{\thefigure}{\thesection.\arabic{figure}}
\renewcommand{\thetable}{\thesection.\arabic{table}}
\renewcommand{\theequation}{\thesection.\arabic{equation}}

\clearpage
\setcounter{page}{1}

{
\centering
\Large
\vspace{1.0em}
\textbf{Supplementary Material}\\
}
\appendix

\vspace{1em}
This supplementary material presents additional experimental results (Section~\ref{sec:additional_qual_results}) and detailed prompts used by each method in the ablation study (Section~\ref{sec:ablation_configurations}).
Qualitative video results can be found on the project page: \href{https://gem-nr.github.io}{https://gem-nr.github.io}.

\section{Additional Experimental Results}
\label{sec:additional_qual_results}
Below, we provide additional experimental results, including evaluation of the 2D editing backbones; analysis of sensitivity to increased disparity, reconstruction errors, and global motion change; and further qualitative examples of edited images and renders.

\subsection{Backbone choice}
There are two types of 2D editing backbones that can be utilized within the GeM-NR framework: one for \textit{anchor editing} and one for \textit{refinement after warping}. For anchor editing, any 2D editor can be used, and in the main paper, we evaluate multiple state-of-the-art models.
\begin{figure}[h]
    \centering
    \caption{\textbf{Runtime breakdown and consistency evaluation on object addition task.} Runtime is shown on a semi-logarithmic scale, excluding anchor editing. The reported consistency metrics are CLIP-based (EC) and epipolar distance-based (mAA). FLUX.2 [klein] 9B offers a good trade-off between runtime and performance.}
    \includegraphics[width=\linewidth, trim={0 0 0 0}, clip]{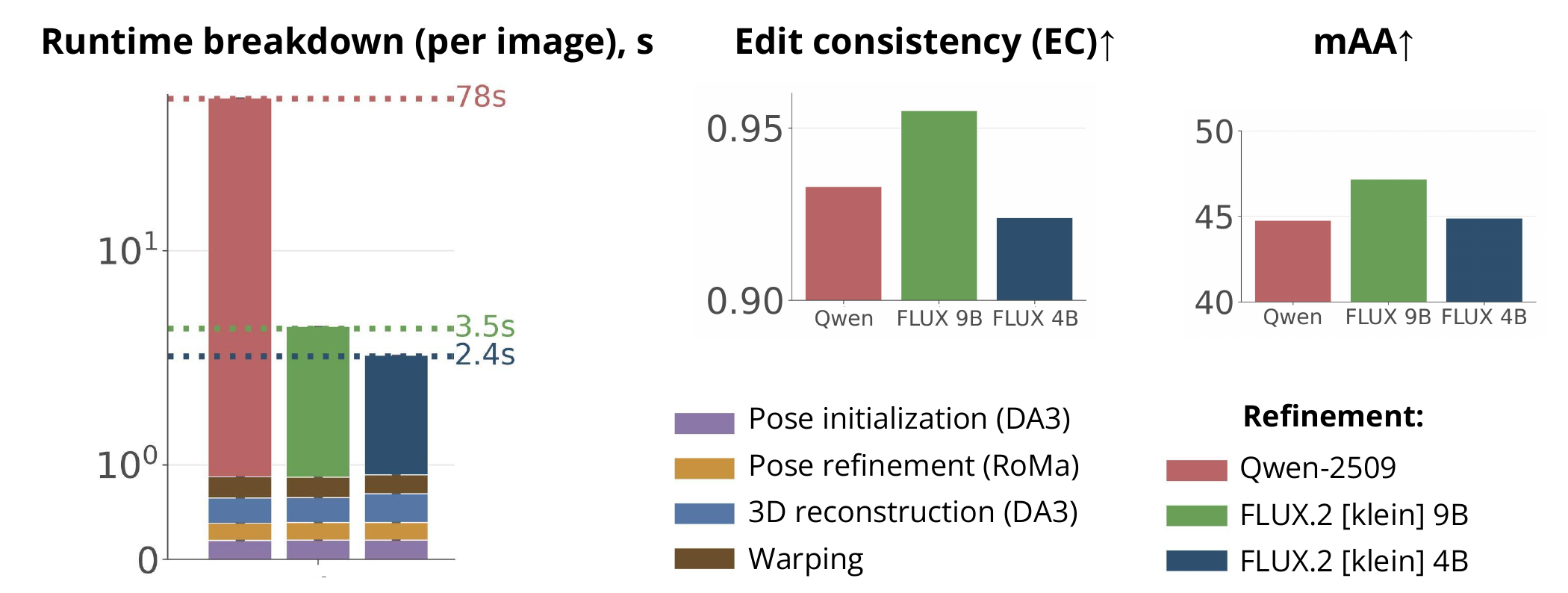}
    \label{runtime_breakdown}
\end{figure}
For refinement after warping, the 2D editor must support multi-reference. We limit our choice to FLUX.2 or Qwen models, which are recent and commonly used methods for this task. We evaluate these models on the object addition task and provide a breakdown of their runtime and performance in Fig.~\ref{runtime_breakdown}. FLUX.2 9B offers a good trade-off between speed and performance, motivating our default choice of FLUX.2 9B as the refiner. We find that GeM-NR performs best when the anchor and refinement editors match (see Table~\ref{tab:MultiViewConsistency}). This is unsurprising: the refiner is more likely to produce an image consistent with the edited anchor when it can reproduce the same edit as the anchor editor.

\subsection{Sensitivity to increased disparity}
\begin{figure}[h]
  \centering
  \caption{\textbf{Effects of increased disparity} on warped/edited images.}
  \scriptsize{\textit{``Turn the bear statue into a magical creature statue''}}\\
  \includegraphics[width=\linewidth]{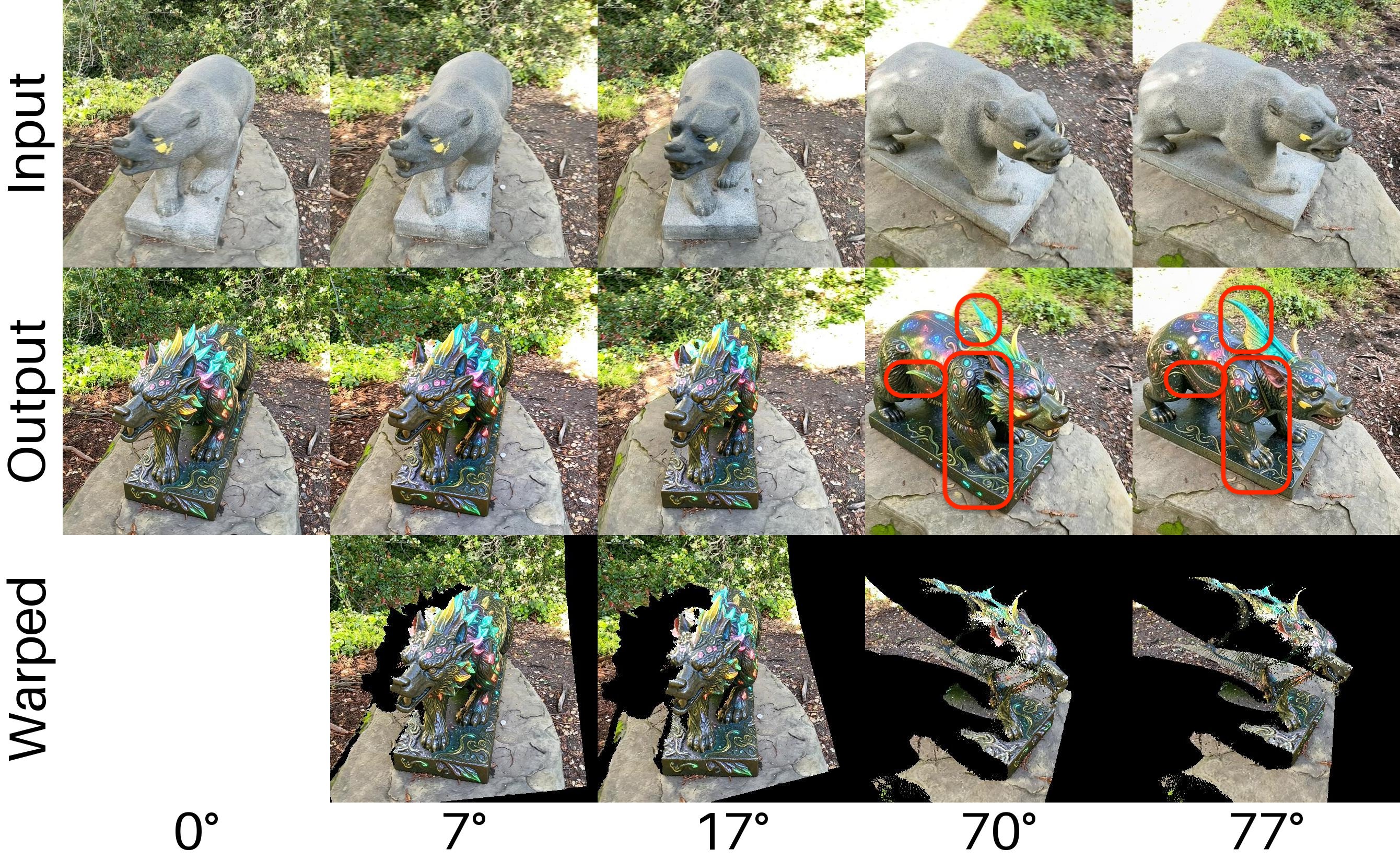}
   \label{warp360}
\end{figure}

Fig.~\ref{warp360} illustrates an example of editing a query image with an increasingly wide baseline relative to the anchor image. This results in larger missing regions in the warped images, providing less reference information for refinement and greater freedom for hallucination. Consequently, query images forming a wide baseline with the anchor image have a higher risk of localized inconsistencies, as illustrated in Fig.~\ref{warp360} (70°–77°). Addressing this is a promising future research direction.

\subsection{Sensitivity to reconstruction errors}
Fig.~\ref{robustness_da3} shows an example in which Depth Anything 3 fails to accurately reconstruct the scene, resulting in flatter facial surfaces. We find that reconstruction errors frequently occur in facial regions. GeM-NR is designed to be modular and can therefore readily accommodate future upgrades to the underlying backbones. However, we also observe that GeM-NR in its current form is often robust to such errors (see Fig.~\ref{robustness_da3}). When the warped region is implausible, FLUX tends to rely less on it and more on the text prompt and input image.

\subsection{Sensitivity to global motion change}
\begin{figure}[h]
\centering
  \begin{minipage}[t]{0.43\linewidth}
  \caption{\textbf{DA3~error}~example giving flatter face surface. FLUX ignores this error.}
    \centering
    \scriptsize\textit{``Make him wear a T-shirt''}\\
    \includegraphics[width=\linewidth]{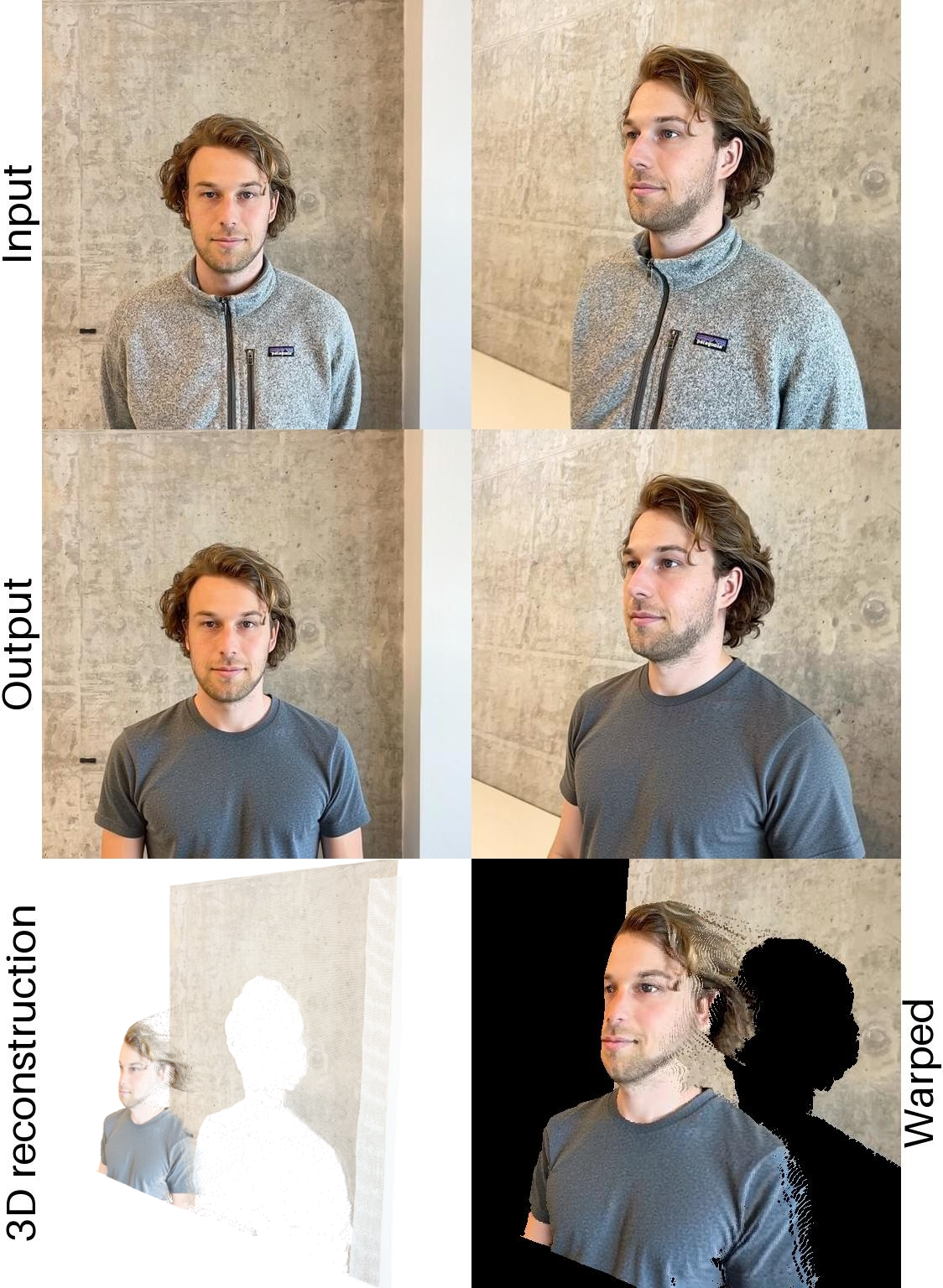}
   \label{robustness_da3}
   \end{minipage}
  ~~~
  \begin{minipage}[t]{0.52\linewidth}
  \caption{\textbf{Global motion change:} results without and with global motion estimation.}
    \centering
    \scriptsize{\textit{``Remove the boxes behind the piano, change the pot to a gray clay pot with the gray clay bottom, \textcolor{blue}{turn the camera slightly to align the horizon and such that the table [...] is not visible}''}}\\
    \includegraphics[width=\linewidth]{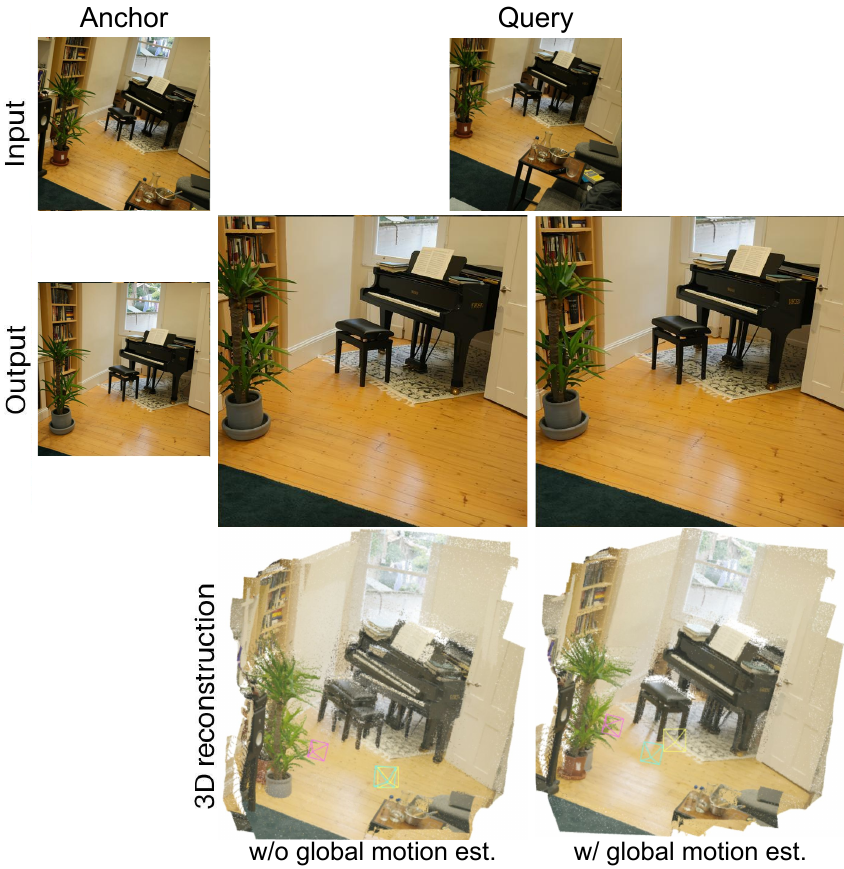}
   \label{robustness}
  \end{minipage}
\end{figure}

While our problem setting naturally assumes unchanged global motion, we tried to violate this assumption in order to assess the sensitivity of GeM-NR to global motion change. We simulate this scenario by instructing the model to move the camera, as shown in Fig.~\ref{robustness}. The resulting edit is visually pleasing but exhibits a small, barely noticeable error, stemming from an expected misalignment between the original and edited point clouds (Fig.~\ref{robustness}, bottom left). A straightforward fix is to incorporate A. $\leftrightarrow$ E.A. relative pose estimation, which ensures improved alignment (see Fig.~\ref{robustness}, bottom right) and greater 3D consistency during editing.

\newcolumntype{M}[1]{>{\centering\arraybackslash}m{#1}}

\subsection{Additional evaluation results}
\begin{figure}[h]
\centering
\caption{\textbf{Edited multi-view images.} GeM-NR gives consistent edits with preserved details, such as the bicycle parts and the patches on the jacket.}
{\scriptsize \textit{`Change the bicycle to an dirt motorbike''}}\\
\includegraphics[width=\textwidth]{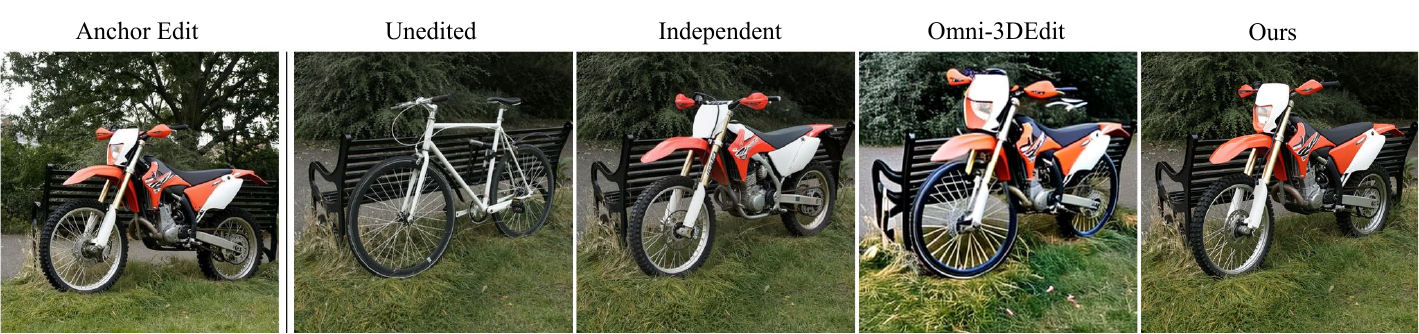}\\
 {\scriptsize \textit{``Give him a leather jacket with different patches''}}\\
\includegraphics[width=\textwidth]{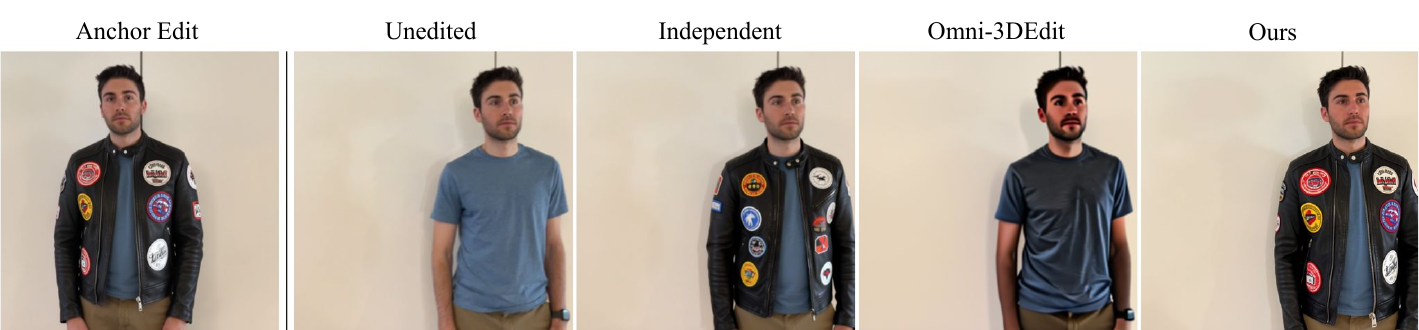}\\
\label{fig:qual10}
\end{figure}

\begin{figure}[h]
\centering
\caption{\textbf{Additional qualitative multi-view editing results with Qwen anchor editor.}}
{\scriptsize\textit{``Make him carry a bag of groceries''}}\\
\includegraphics[width=1\textwidth]{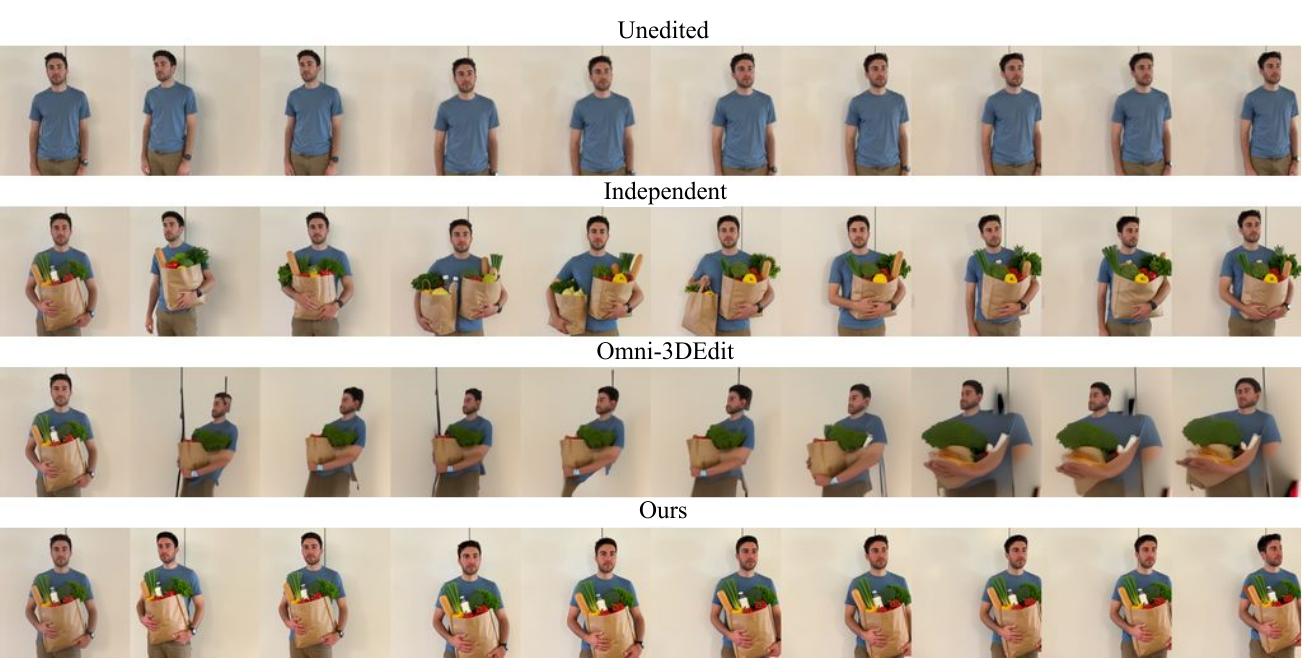}\\[4pt]
\label{fig:qual1}
\end{figure}

\begin{figure}[h]
\centering
\caption{\textbf{Additional qualitative multi-view editing results with Qwen anchor editor.}}
\vspace{-3pt}
{\scriptsize\textit{``Change bear statue to a sitting dog''}}\\
\includegraphics[width=1\textwidth]{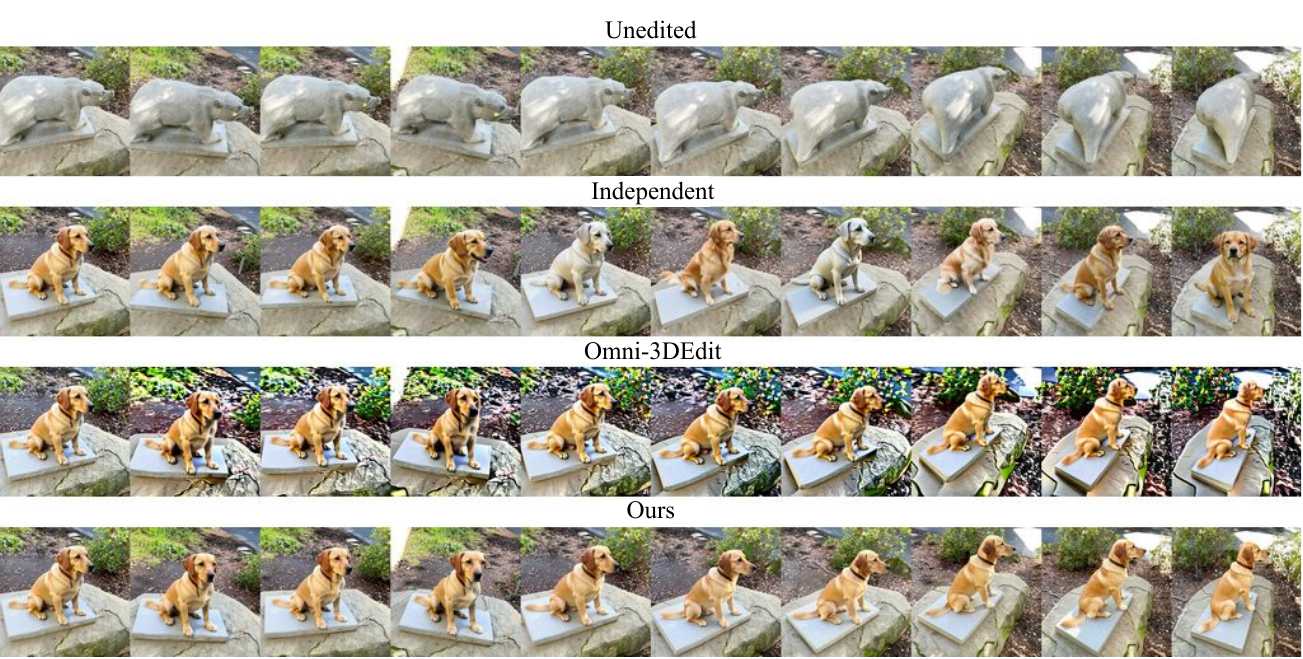}
\vspace{-18pt}
\label{fig:qual2}
\end{figure}

In Table~\ref{tab:CLIPRenders2}, we report text alignment metrics for renderings of the edited 3DGS when FLUX.2 [klein] is used to edit the initial anchor image. The results are comparable to those obtained when using Qwen to edit the anchor image (see Table~\ref{tab:eval_pairs}). Additional qualitative examples of the multi-view consistent editing are provided in Figures~\ref{fig:qual10}, \ref{fig:qual1}, \ref{fig:qual2}.
\begin{table}[h]
\centering
\footnotesize
\caption{\textbf{Text alignment (TA$\uparrow$ / TA dir$\uparrow$) of the renders from 3DGS of edited scene with FLUX.2 [klein] anchor backbone}. GeM-NR clearly improves over Omni-3DEdit. It especially improves for the general nonrigid editis and object addition, where both Independent and Omni-3DEdit struggle.}
\resizebox{\textwidth}{!}{
\begin{tabular}{l|ccccc}
\toprule
& All Types
& General Nonrigid
& Object Addition
& Object Removal
& Appearance Change \\
\midrule
\textit{Unedited}
& 0.201 / -
& 0.194 / -
& 0.201 / -
& 0.200 / -
& 0.206 / - \\
Independent
& 0.253 / 0.192
& 0.254 / 0.227
& 0.243 / 0.169
& \textbf{0.213} / \textbf{0.170}
& 0.271 / 0.187 \\
Omni-3DEdit
& 0.233 / 0.131
& 0.244 / 0.185
& 0.240 / 0.140
& 0.206 / 0.114
& 0.233 / 0.101 \\
\textbf{Ours}
& \textbf{0.259} / \textbf{0.205}
& \textbf{0.270} / \textbf{0.269}
& \textbf{0.253} / \textbf{0.176}
& 0.211 / 0.163
& \textbf{0.272} / \textbf{0.190} \\
\bottomrule
\end{tabular}
}
\label{tab:CLIPRenders2}
\end{table}

\vfill
\newpage
\clearpage
\setcounter{table}{0}
\section{Ablation Details}
\label{sec:ablation_configurations}
Table~\ref{tab:ablations_simple} presents the detailed instructions used by each method in the ablation study. These instructions are used during the refinement stage. GeM-NR is highlighted in gray.

\begin{table}
\centering
\footnotesize
\caption{\textbf{Instructions corresponding to each method in the ablation study.}}
\begin{tabular}{HccccM{6cm}}
\toprule
\textbf{} &
Reference text &
\multicolumn{3}{c}{Reference image(s)} &
\multicolumn{1}{c}{\multirow{2}{*}{Instruction}}
\\
\cmidrule(lr){3-5}
 &
\shortstack{Input\\edit} &
\shortstack{Original\\query} &
\shortstack{Edited\\anchor} &
\shortstack{Depth-\\warp} &
 \\
\midrule
\multicolumn{6}{c}{\textbf{Simultaneous pair editing}} \\
\multicolumn{6}{l}{\textit{Concatenate both inputs}} \\
ConcatenatedPair & \checkmark & N/A & N/A & N/A & \scriptsize\texttt{``Edit these two concatenated images in a consistent way''}\\
\midrule
\multicolumn{6}{c}{\textbf{Editing one conditioned on another}} \\
\multicolumn{6}{l}{\textit{Original edit text prompt preserved; no warp}} \\
Independent      & \checkmark & \checkmark & \ding{55} & \ding{55}  & \scriptsize\texttt{\textbf{T}} \\
Baseline         & \checkmark & \checkmark & \checkmark & \ding{55} & \scriptsize\texttt{``Edit the first image as shown in the second image''} \\
\multicolumn{6}{l}{\textit{Original edit text prompt ignored; with warp}} \\
WarpInpaint      & \ding{55} & \ding{55}  & \ding{55}  & \checkmark & \scriptsize\texttt{``Inpaint this image''} \\
WarpInpaint2     & \ding{55} & \checkmark & \ding{55}  & \checkmark & \scriptsize\texttt{``Edit this image as shown in the second image. Inpaint the second image''} \\
WarpInpaint3     & \ding{55} & \ding{55}  & \checkmark & \checkmark & \scriptsize\texttt{``Inpaint the first image. Use the second image as a guidance on how to inpaint''} \\
NoInitialPrompt  & \ding{55} & \checkmark & \checkmark & \checkmark &  \scriptsize\texttt{``Edit the first image in the same way as shown in the second image. The suggested appearance is in the third image. Stick to this change, but refine it to keep consistency with respect to the first image''} \\
\multicolumn{6}{l}{\textit{Original edit text prompt preserved; with warp}} \\
WAV3D\_WithAnchor       & \checkmark & \checkmark & \checkmark & \checkmark & \scriptsize\texttt{concat(\textbf{T}, ``The suggested appearance is in the second image. Stick to this change, but refine it to keep consistency with respect to the first image. For reference, the same edit at a different viewpoint is provided in the third image'')} \\
\rowcolor{gray!30} WAV3DSingleCondPipeline & \checkmark & \checkmark & \ding{55} & \checkmark & \scriptsize\texttt{concat(\textbf{T}, ``The suggested appearance is in the second image. Stick to this change, but refine it to keep consistency with respect to the first image'')} \\
\bottomrule
\end{tabular}
\label{tab:ablations_simple}
\end{table}

\end{document}